\documentclass[10pt]{article}
\expandafter\let\csname equation*\endcsname\relax
\expandafter\let\csname endequation*\endcsname\relax
\usepackage{bm}
\usepackage{pifont}
\usepackage{enumerate}
\usepackage[top=1in, bottom=1in, left=1in, right=1in]{geometry}
\usepackage[dvipsnames]{xcolor}
\usepackage{mathrsfs}
\usepackage{amsfonts}
\usepackage{multirow}
\usepackage{upgreek}
\usepackage{nicefrac}  
\usepackage{amssymb,dsfont}
\usepackage{caption,comment}
\usepackage{blkarray}
\usepackage{amsmath}
\usepackage{float}
\usepackage{footnote}
\usepackage{amssymb,amsthm}
\usepackage[toc,page]{appendix}
\usepackage{graphicx,afterpage}
\usepackage{epstopdf}
\usepackage[normalem]{ulem}
\usepackage{algorithm}
\usepackage{algorithmic}
\usepackage{xspace}
\usepackage{enumitem}
\usepackage{authblk}
\usepackage[pdftex,colorlinks=true]{hyperref}
\hypersetup{CJKbookmarks,%
	bookmarksnumbered,%
	colorlinks,%
	linkcolor=black,%
	citecolor=black,%
	plainpages=false,%
	pdfstartview=FitH}
\numberwithin{equation}{section}
\numberwithin{figure}{section}

\newcommand\tabcaption{\def\@captype{table}\caption}

\newtheorem{thm}{Theorem}[section]

\newtheorem{rem}[thm]{Remark}

\newtheorem{lemma}[thm]{Lemma}
\newtheorem{proposition}[thm]{Proposition}
\newtheorem{assumption}{Assumption}

\def\real{\mathbb{R}}

\newcommand{\cP}{\mathcal P}

\newcommand{\cT}{\mathcal T}

\renewcommand{\d}{\,\mathrm{d}}
\newcommand{\R}{\mathbb R}

\newcommand{\KL}{\mathrm{KL}}

\newcommand{\Cov}{\mathrm{Cov}}

\newcommand{\supp}{\mathrm{supp}}

\def\F{\textrm{FR}}
\def\G{\textrm{G}}
\def\W{\textrm{W}}

\usepackage{tikz}
\usetikzlibrary{positioning}
\title{An operator splitting analysis of Wasserstein--Fisher--Rao \\gradient flows}
\date{}

\author[1]{Francesca Romana Crucinio\thanks{ \href{mailto:francescaromana.crucinio@unito.it}{francescaromana.crucinio@unito.it}
    F.R.C. gratefully acknowledges the ``de Castro" Statistics Initative at the \textit{Collegio Carlo Alberto} and the \textit{Fondazione Franca e Diego de Castro}. F.R.C. is supported by the Gruppo
Nazionale per l'Analisi Matematica, la Probabilità e le loro Applicazioni (GNAMPA-INdAM).}}
\author[2]{Sahani Pathiraja\thanks{  \href{mailto:s.pathiraja@unsw.edu.au}{s.pathiraja@unsw.edu.au} S.P. gratefully acknowledges funding from UNSW Faculty of Science Research Grant and the Eva Mayr Stihl Foundation.}}
\date{ }

\affil[1]{ESOMAS, University of Turin, Italy \& Collegio Carlo Alberto, Turin, Italy}
\affil[2]{School of Mathematics \& Statistics, UNSW Sydney, Australia}

\begin{document}

    \maketitle
\vspace{-0.7cm}

\begin{abstract}
  Wasserstein-Fisher-Rao (WFR) gradient flows have been recently proposed as a powerful sampling tool that combines the advantages of pure Wasserstein (W) and pure Fisher-Rao (FR) gradient flows. Existing algorithmic developments implicitly make use of operator splitting techniques to numerically approximate the WFR partial differential equation, whereby the W flow is evaluated over a given step size and then the FR flow (or vice versa).  This work investigates the impact of the order in which the W and FR operator are evaluated and aims to provide a quantitative analysis.  We demonstrate a counterintuitive result that with a judicious choice of step size and operator ordering, the split scheme can converge to the target distribution faster than the exact WFR flow (in terms of model time). We obtain variational formulae describing the evolution over one time step of both splitting schemes and investigate in which settings the W-FR split should be preferred to the FR-W split.
\end{abstract}

	\tableofcontents

\section{Introduction}
\label{sec:intro}
We consider the task of generating samples from a target probability distribution known up to a normalisation constant with density $\pi(x) \propto e^{-V_\pi(x)}, \enskip x \in \mathbb{R}^d$.  Despite the conceptual simplicity of this task, its efficient implementation when $V_\pi$ is multi-modal, the underlying space is high dimensional and/or modes are separated by large distances remains challenging.   Given the broad application of sampling to (Bayesian) statistics and statistical machine learning, several avenues have been investigated to derive efficient and scalable algorithms.

A natural way to formulate this task is via gradient flows, which can be seen as optimisation of a functional measuring the dissimilarity to $\pi$, typically the Kullback--Leibler (KL) divergence \cite{wibisono2018sampling, chen2023sampling, crucinio2025note}.
This formulation yields considerable freedom in the design of sampling algorithms; arguably the most well-known being the so-called Wasserstein Gradient flow, hereafter W flow \cite{jordan1998variational}.  
It is well-known that when $\pi$ satisfies a Log-Sobolev Inequality (LSI), the W flow converges exponentially fast to $\pi$, with rate depending on the Log-Sobolev constant.  This highlights an inherent limitation of W flows, namely that when the LSI constant is large (e.g. as is typically the case in multi-modal densities with well separated modes), the convergence rate can be prohibitively slow \cite{schlichting2019poincare}.

Recent research efforts have instead considered gradient flows in the Fisher--Rao geometry (FR flow).  These are well known in the biological literature as describing the macroscopic properties of a population with varying traits or species, and are sometimes referred to as birth-death or replicator dynamics \cite{kimura_stochastic_1965,schuster1983replicator, Cressman2006}.  It is known that FR flows are intimately connected to mirror descent \cite{chopin2023connection}, stochastic filtering \cite{akyildiz2017probabilistic, halder_gradient_2017, Pathiraja2024, DelMoral1997} and sequential Monte Carlo \cite{us}.  They have also recently been exploited to develop sampling algorithms \cite{Nusken2024, maurais2024sampling, Wang2024, chen2023sampling, lu2023birth, Lu2019}, most notably due to the fact that it is possible to achieve convergence rates independent of the properties of $V_\pi$ \cite{carrillo_fisher-rao_2024, Lu2019}. However, the development of stable and efficient numerical approximations of pure FR flows remains a challenge. 

In this work, we consider the so-called Wasserstein--Fisher--Rao (WFR) gradient flow, wherein the metric is given by the direct sum of the Wasserstein and Fisher-Rao metrics.
We refer the reader to \cite{liero_optimal_2018} for a rigorous treatment of this metric and its corresponding gradient flow. 
WFR gradient flows combine the diffusive behaviour of W flows with the birth-death or reactive properties of FR flows and enjoy better convergence properties than both W and FR alone \cite{Lu2019, chen2023sampling, us_bernoulli}.  They have a natural interpretation as combining `exploration' or `mutation' to provide new particles (W flow) with `selection' where particles that are a poor fit to the target are killed \cite{Pathiraja2024}. Research into WFR flows is active; several open questions and areas of investigation remain both in terms of algorithmic development and theoretical investigation.

The WFR partial differential equation (PDE)  describes the continuous-time dynamics underlying several recent sampling methods \cite{us, lu2023birth, Lu2019}. Because it is a sum of Wasserstein and Fisher--Rao operators, it can be approximated naturally by Lie--Trotter splitting: the W and FR subproblems are solved successively over a chosen step size.  Such operator splitting strategies have been used to approximate Langevin and Kinetic Langevin dynamics \cite{abdulle2015long, duncan2017nonreversible, chada2023unbiased, leimkuhler2013rational, bou2010long}. This approach is particularly natural here because the W flow contains diffusion and transport, whereas the FR flow is a reaction equation. However, the effect of the ordering of the two subflows has not been quantified.

We show that a carefully chosen splitting can exhibit faster convergence to the target than the exact dynamics for suitable targets, initial laws, and step sizes.
This effect is due to an ordering-dependent splitting error (which preserves $\pi$) and does not require additional elementary W/FR evaluations when cost is measured at fixed substep size. This suggests designing numerical methods for the split dynamics themselves; for example, one may group a sequence of W steps followed by a sequence of FR steps, or reverse the order, rather than alternating the two operators at every step.  As motivating evidence, consider Figure \ref{fig:Gauss10D}(right) which shows the significant improvement in KL simply by modifying the ordering of a sequence of W and FR steps.
In this work, we focus on exact splitting, i.e.
we do not consider numerical schemes to approximately solve the W and FR flow. 
There are a myriad schemes to numerically approximate both W \cite{NEURIPS2019_6a8018b3_salim, salim2020wasserstein, mou2021high} and FR \cite{nusken2024transport, maurais2024sampling, Wang2024, korba2022adaptive, chen2023sampling} flows, we leave their detailed investigation to future work. 

Our main contributions are as follows:
\begin{itemize}
    \item Variational formulae (PDEs) that quantify the impact of operator splitting as function of the step size (Section~\ref{sec:splitting}). 
    \item We provide as motivation, a precise quantitative characterisation of the effects of operator ordering in the multivariate Gaussian case (Section~\ref{sec:gaussian}). 
    \item A quantitative description of sufficient conditions for log-concave targets under which the split scheme converges to $\pi$ faster than the exact WFR flow.
\end{itemize}

\paragraph*{Notation}
We define some notation that will be used throughout the manuscript.
For all differentiable functions $f$ we denote the gradient by $\nabla f$. Furthermore, if $f$ is twice differentiable we
denote by $\nabla^2f$ its Hessian and by $\Delta f$ its Laplacian. 
 We denote by $\cP(\real^d)$ the set of probability measures over
$\mathcal{B}(\real^d)$, and endow this space with the topology of weak convergence. 
We denote by $\cP_2^{ac}(\real^d)$ the manifold of absolutely continuous probability measures on $\real^d$ with finite second moment. Every $p\in \cP_2^{ac}(\real^d)$ will be identified with its (Lebesgue) density $p(x)$. At a $p\in \cP_2^{ac}(\real^d)$ such that $p(x) > 0$ everywhere, the tangent space of $\cP_2^{ac}(\real^d)$ is given by $\cT  = \{ \sigma\in C^\infty(\R^d): \int \sigma(x)\d x = 0\}$. If $p$ has support $\supp(p)\subset \R^d$, then the tangent space is $\cT  = \{ \sigma\in C^\infty(\R^d): \sigma|_{\supp(p)^C} = 0, \int \sigma(x)\d x = 0\}$.  Throughout this manuscript we denote the target by $\pi$ and assume $\pi(x) \propto e^{-V_\pi(x)}$. The initial distribution of each flow is denoted by $\mu_0$.  
The Kullback--Leibler divergence is defined for $\nu,\mu$ admitting a density w.r.t. Lebesgue as $\KL(\nu||\mu)=\int \log(\nicefrac{\nu(x)}{\mu(x)}) \nu(x)dx$.

\section{Operator Splitting for the WFR PDE}
\label{sec:splitting}

The gradient flow of $\KL(\mu || \pi)$ w.r.t. the geometry induced by the Wasserstein--Fisher--Rao distance is given by the following PDE \cite[Theorem 3.1]{Lu2019}
\begin{align}
    \label{eq:WFRpde}
    \partial_t \mu_t &= f_\W(\mu_t)+f_{\F}(\mu_t),\\
    f_\W(\mu) &:= \nabla \cdot (\mu \nabla g(\mu)), \label{eq:winfflow}\\
    f_{\F}(\mu) &:= -\mu \left(g(\mu) -  \mathbb{E}_\mu \left[ g(\mu) \right]  \right), \label{eq:infFR}\\
    \label{eq:geqn}
    g(\mu) &:= \log \frac{\mu}{\pi}.
\end{align}
and $f_\W$ and $f_{\F}$ correspond to the Wasserstein and Fisher--Rao operators respectively, and takes the form of a advection--diffusion--reaction equation. 
This PDE is a combination of the Wasserstein (W) flow and of the Fisher--Rao (FR) flow and enjoys improved convergence properties compared to both W and FR alone \cite{Lu2019, chen2023sampling, us_bernoulli}.

Throughout this manuscript, we focus on time discretisation of \eqref{eq:WFRpde} via operator splitting under the assumption that the operators are evaluated exactly over a single time step, i.e. without further discretisations using either deterministic or Monte Carlo type approximations.  Our motivation for doing so is to quantify biases due solely to splitting error, thereby giving rise to a new PDE which can potentially yield faster (continuous time) convergence rates than the original WFR PDE.  This opens the door for the development and study of numerical schemes whose underlying dynamics correspond to the split WFR flow rather than its fully continuous time counterpart.  
In particular, we assume that the solution operator of the W flow over a time step $\gamma$ applied to measure $v$, $S_\W(\gamma, v)(x)$,
can be evaluated exactly.
Similarly, we assume the solution operator of the FR flow given by (e.g. \cite[App. B.1]{chen2023sampling}, \cite[Eq. (16)]{lu2023birth})
\begin{align}
\label{eq:fr_semigroup}
   S_\F(\gamma, v)(x)\propto \pi(x)^{1 - e^{-\gamma}} v(x)^{e^{-\gamma}},
\end{align}
can also be evaluated exactly. 
 A commonly used first order accurate splitting method is sequential splitting or Lie-Trotter splitting (see, e.g., \cite{blanes2024splitting} for a recent review), where each operator is solved over a step size in an alternating manner. More specifically, consider a time-discretisation $t_0 = 0 < t_1 < t_2 \cdots < t_M = T$ with $t_{n+1} - t_n = \gamma \enskip \forall \enskip i = 1, 2, \dots M$.  A sequential splitting applied to \eqref{eq:WFRpde} takes the form 
\begin{align}
\tag{W-FR}  \label{eq:sequential_split}
    \hat{\nu}_i(x; \gamma) &= S_\W(\gamma,  \nu_{i-1})(x) \rightarrow \nu_i(x; \gamma) = S_\F(\gamma,\hat{\nu}_{i})(x), \quad i = 2, 3, \dots  
\end{align}
with $\nu_0 = \mu_0$ and
where $S_\F(\gamma, v)$ and $S_\W(\gamma, v)$ denote the solution operators corresponding to $f_{\F}$ and $f_\W$ respectively, acting on $v$ over time interval of size $\gamma$. We refer to~\eqref{eq:sequential_split} as Wasserstein then Fisher--Rao scheme.  We note that the step size $\gamma$ for the W and FR components need not be the same.   
It is known that $ \nu_n \rightarrow \mu_{n\gamma}, \enskip \gamma \rightarrow 0$ uniformly in time at rate $\mathcal{O} (\gamma)$ under mild conditions on $f_\F$, $f_\W$ and $\pi$ (e.g. \cite{Hundsdorfer2003}).  

The splitting scheme in \eqref{eq:sequential_split} corresponds to evaluating the W solution operator first. A valid alternative sequential splitting scheme to \eqref{eq:sequential_split} involves instead evaluating the FR solution operator first, i.e.
\begin{align}
        \hat{\eta}_i(x; \gamma) &= S_\F(\gamma,  \eta_{i-1})(x) \rightarrow \eta_i(x; \gamma) = S_\W(\gamma,\hat{\eta}_{i})(x), \quad i = 1, 2, 3, \dots  
        \tag{FR-W} \label{eq:sequential_splitFRthenW}
\end{align}
We refer to~\eqref{eq:sequential_splitFRthenW} as the Fisher--Rao then Wasserstein scheme. 
When the operators $S_\F$ and $S_\W$ do not commute, it is known that $\eta_i(x;\gamma) \neq \nu_i(x;\gamma)$ for $\gamma > 0$ and $i = 1, 2, 3, \dots$ \cite{Hundsdorfer2003}.  It is expected that for most target distributions,  $S_\F$ and $S_\W$ will not commute; indeed this can be straightforwardly verified even in the multivariate Gaussian case as we show in Section~\ref{sec:gaussian}.

\begin{rem}
    One could also apply a splitting scheme for pure W to split the advection and diffusion parts (leading to proximal samplers, see e.g. \cite[Section 4.2]{wibisono2018sampling}). Similarly, one could also apply a splitting scheme to the pure FR flow by identifying an exploration step and an exploitation step as in \cite[Section 4]{chen_efficient_2024}. We leave analysis of these alternative splitting schemes for future work.
\end{rem}

Throughout we rely on the following assumptions on the target distribution $\pi$ and the distribution at which the WFR PDE~\eqref{eq:WFRpde} is initialised  $\mu_0$ in developing analytic results.  
\begin{assumption}
\label{ass:lsi} The following conditions on the initial and target distribution hold.
\begin{enumerate}[label=(\alph*)]
\item \label{ass:pi} $\pi(x) \propto e^{-V_\pi(x)}$ where $V_\pi\in C^2(\real^d)$ with bounded Hessian, and satisfies a Log-Sobolev inequality (LSI): for all $\mu\in \cP_2^{ac}(\real^d)$
\begin{align}
    \label{eq:lsi}
    \KL(\mu||\pi)\leq \frac{\lambda_\pi}{2}\int \mu \left\lvert\nabla\log \frac{\mu}{\pi}\right\rvert^2:= \frac{\lambda_\pi}{2}\mathcal{I}(\mu||\pi),
\end{align}
where $\lambda_\pi<\infty$ denotes the Log-Sobolev constant of $\pi$ and is $\mathcal{I}(\mu||\pi) $ the relative Fisher information between $\mu$ and $\pi$;
\item \label{ass:mu0} $\mu_0(x) \propto e^{-V_0(x)}$ where $V_0\in C^2(\real^d)$ with bounded Hessian, satisfies an LSI with constant $\lambda_0<\infty$ and $\mu_0\ll \pi$.
\end{enumerate}
\end{assumption}
Assumption~\ref{ass:lsi}\ref{ass:pi} implies that the Wasserstein flow targeting $\pi$, $\partial_t \mu_t = f_\W(\mu_t)$, is well defined and the solution of the corresponding PDE is absolutely continuous w.r.t. $\pi$ for all $t\geq 0$. The bounded Hessian requirement implies that $\nabla V_\pi$ is Lipschitz continuous, we denote the Lipschitz constant by $L_\pi$.
Assumption~\ref{ass:lsi}\ref{ass:mu0} is mild since $\mu_0$ is often user-chosen and guarantees that $\nabla V_0$ is Lipschitz continuous with constant $L_0$. 

We further assume that the FR PDE $\partial_t \mu_t = f_\F(\mu_t)$ and WFR PDE $\partial_t \mu_t = f_\W(\mu_t)+f_\F(\mu_t)$ are well-posed and that their solutions, where they exist, are absolutely continuous w.r.t. $\pi$. These assumptions guarantee that the solution operators $S_\W, S_\F$ are well-defined.

In the remainder of the manuscript, we use the notation $\nu_i(x)$ and $\eta_i(x)$ in place of $\nu_i(x;\gamma)$ and $\eta_i(x; \gamma)$ respectively as we assume a step size of $\gamma$ throughout.  In the next section, we investigate the behaviour of these quantities as $\gamma$ varies. 


We begin by analysing the splitting scheme in~\eqref{eq:sequential_split}.  In particular, we obtain a PDE which describes the evolution of the splitting solution over time interval $[0, \gamma]$.  This can equivalently be seen as describing how $\nu_1$ varies with respect to the step size.   
The advantage of this characterisation is that we can identify a perturbation to the exact dynamics and more easily characterise how splitting biases modify convergence speed.  The traditional route to analysing decay of $\KL$ of the WFR dynamics in e.g. \cite[Proposition 3.1]{us},  \cite[Remark 2.6]{lu2023birth}, \cite[Appendix B]{Lu2019} is not suited to the splitting schemes as it does not allow us to distinguish 
the order in which the W and FR solution operators are applied.

The following proposition quantifies the variation of $\nu_{\gamma}$ with respect to the step size $\gamma$ using standard variational calculus tools. Its proof can be found in Appendix~\ref{app:proofseqWFR}. 

\begin{proposition}
\label{lem:seqsplitpdeWthenFR}
    \textbf{Sequential split PDE: W-FR}  Let us denote  $\nu_1 = \nu_1(x;\gamma)$ and let Assumption~\ref{ass:lsi} hold. The variation of one sequential split step in the order W-FR \eqref{eq:sequential_split} of size $\gamma$ corresponds to the PDE 
    \begin{align}
    \label{eq:seqsplitpdeWthenFR}
         \partial_\gamma \nu_1 &= f_\W(\nu_1) + f_\F(\nu_1) + (e^\gamma - 1)f_{\textrm{P}}(\nu_1) \\
         \label{eq:perturbation}
 f_{\textrm{P}}(\nu_1) &:= \nu_1\left( \left|\nabla g(\nu_1) \right|^2 - \mathbb{E}_{\nu_1} \left[ \left|\nabla g(\nu_1) \right|^2 \right]  \right),
    \end{align}   
    where $g$ is as defined in \eqref{eq:geqn}. 
\end{proposition}

The above proposition shows that splitting introduces a perturbation of Fisher-Rao structure to the WFR PDE, whose contribution is modulated by an additional factor $(e^\gamma -1)$ which increases monotonically with step size $\gamma$.  
Somewhat remarkably, this simple splitting procedure is sufficient to generate an acceleration under certain conditions, as we will demonstrate in Section~\ref{sec:gaussian} for the Gaussian case and in Section~\ref{sec:convlogconc} for the log-concave case. 


The reverse scheme described in \eqref{eq:sequential_splitFRthenW} requires a more involved analysis than in Proposition \ref{lem:seqsplitpdeWthenFR}, due to the lack of invertibility of $S_\W$ in general. 
We instead rely on techniques used in developing Alekseev-Groebner type formulae, in particular, the formal Lie calculus, as is common in the analysis of operator splitting schemes \cite{descombes_lietrotter_2013, Hundsdorfer2003}. 
We have the next proposition for the FR-W scheme.  Its proof can be found in Appendix~\ref{app:proofseqFRW}.

\begin{proposition}
    \label{lem:seqsplitpdeFRthenW}
    \textbf{Sequential split PDE: FR-W.} Let us denote  $\eta_1 = \eta_1(x;\gamma)$ and let Assumption~\ref{ass:lsi} hold. The variation of one sequential split step of size $\gamma$ in the order FR-W \eqref{eq:sequential_splitFRthenW} corresponds to the PDE 
    \begin{align}
    \label{eq:PDE_splitting2}
  \partial_\gamma \eta_1 = f_\W(\eta_1) + f_\F(\eta_1) + \int_0^\gamma S_\W\left(\gamma - \tau, -f_{\textrm{P}}(\eta_1(x; \tau))  \right) d\tau,
\end{align} 
where $f_{\textrm{P}}$ is as defined in \eqref{eq:perturbation}. Formally, the following representation in terms of a Lie expansion holds,
    \begin{align}
        \label{eq:seqsplitpdeFRthenW}
        \partial_\gamma \eta_1 = f_\W(\eta_1) + f_\F(\eta_1) +\sum_{k=1}^\infty \frac{\gamma^k}{k!}\eta_1 \left(g_k(\eta_1) - \mathbb{E}_{\eta_1}[g_k]  \right), 
    \end{align}
    where
    \begin{align*}
        g_0(\eta_1) & =  -g(\eta_1)=-  \log \frac{\eta_1}{\pi},\\
        g_k(\eta_1) &= -g_{k-1}'f_\W(\eta_1) + \nabla \log \frac{\eta_1}{\pi} \cdot \nabla g_{k-1}(\eta_1) + \frac{1}{\eta_1} \nabla \cdot (\eta_1 \nabla g_{k-1}(\eta_1)), \quad \text{for} \enskip k = 1, 2, 3, \dots
    \end{align*}
    and $g_{k-1}'f_\W(\eta_1)$ denotes the  directional derivative of $g_{k-1}$ at $\eta_1$ in the direction $f_\W(\eta_1)$. 
    
\end{proposition}

As Proposition~\ref{lem:seqsplitpdeFRthenW} shows, the evolution of the splitting solution obtained with FR-W does not correspond to that obtained in Proposition~\ref{lem:seqsplitpdeWthenFR} for W-FR. 
In the multivariate Gaussian setting, i.e. with $\mu_0(x) = \mathcal{N}(x; m_0, C_0)$ and $\pi(x) = \mathcal{N}(x; m_\pi, C_\pi)$, a more interpretable PDE reminiscent of \eqref{eq:seqsplitpdeWthenFR} can be obtained by direct evaluation of \eqref{eq:seqsplitpdeFRthenW} as 
  \begin{align}
    \label{eq:seqsplitpdeFRthenWGAUSS}
        \partial_\gamma \eta_1 &= f_\W(\eta_1) + f_\F(\eta_1)    -\eta_1 \left( g_{\text{FRW}}(\eta_1) - \mathbb{E}_{\eta_1}[g_{\text{FRW}}(\eta_1)] \right), \\
        g_{\text{FRW}}(\eta) &:=   \nabla g(\eta)^\top \frac{C_\pi }{2}\left(e^{2\gamma C_\pi^{-1}} - I  \right)\nabla g(\eta).\notag
\end{align}  
We investigate the multivariate Gaussian case in further detail in the next section.

\section{Splitting \& convergence speed: multivariate Gaussian case}
\label{sec:gaussian}

The multivariate Gaussian case offers a tractable setting in which to obtain quantitative results on the impact of operator order in splitting of the WFR flow.  More specifically, consider the case $\mu_0(x) = \mathcal{N}(x; m_0, C_0)$ and $\pi(x) = \mathcal{N}(x; m_\pi, C_\pi )$ with $C_\pi \neq C_0$.   It can be readily shown that the solution of the WFR PDE~\eqref{eq:WFRpde} at any $t > 0$ is also Gaussian $\mu_t(x) = \mathcal{N}(x; m_t, C_t)$ with moments given by 
\begin{align}
\label{eq:WFRGaussexactCt}
    C_t &= C_\pi  + (e^{\Gamma t}[(C_0 - C_\pi )^{-1} + (2I + C_\pi )^{-1}]e^{\Gamma t} -(2I + C_\pi )^{-1})^{-1},\\
        \label{eq:WFRGaussexactmt}
    m_t &= m_\pi +  (C_t - C_\pi ) e^{t C_\pi^{-1}}(C_0 - C_\pi )^{-1}(m_0 - m_\pi),
\end{align}
where $\Gamma := C_\pi^{-1} + \frac{1}{2}I$.  A derivation of \eqref{eq:WFRGaussexactCt}-\eqref{eq:WFRGaussexactmt} can be found in Lemma~\ref{lem:analsolngaussian} (letting $M_t = 0$), which extends the derivation of $C_t^{-1}$ provided in \cite{liero2025evolution}.  To the best of our knowledge, the analytic solution for $m_t$ in \eqref{eq:WFRGaussexactmt} has not yet appeared in the literature.  

Similarly, the splitting schemes yield Gaussian solutions at each time, as it is well-known that both W and FR flows preserve Gaussianity. 
We denote by $b_{n}, Q_{n}$ the mean and covariance respectively at step $n$ of the W-FR split scheme with step size $\gamma$ and $a_{n}, P_{n}$ the mean and covariance at step $n$ of the FR-W split. 

The similarity in structure of the sequential split PDEs \eqref{eq:seqsplitpdeWthenFR} and \eqref{eq:seqsplitpdeFRthenWGAUSS} in the Gaussian setting also allows for obtaining analytic solutions of the mean and covariance in a similar fashion.  An application of Lemma~\ref{lem:analsolngaussian} with $\gamma$ in place of $t$ and setting $M_\gamma = (e^\gamma - 1)I$, gives
\begin{align}
    \label{eq:QgammaWthenFR}
    Q_1 &= C_\pi  +  (e^{\Gamma \gamma} (C_0 - C_\pi )^{-1} e^{\Gamma \gamma} + (e^\gamma - 1)C_\pi^{-1})^{-1},\\
    \label{eq:bgammaWthenFR}
    b_1 &=  m_\pi +  (Q_1 - C_\pi ) e^{\gamma C_\pi^{-1}}(C_0 - C_\pi )^{-1}(m_0 - m_\pi).
\end{align}
for the mean $b_1$ and covariance $Q_1$ of a single step of W-FR of size $\gamma$.  Similarly, setting  $M_\gamma = -\frac{1}{2} C_\pi (e^{2\gamma C_\pi^{-1}} - I)$ in Lemma~\ref{lem:analsolngaussian} yields 
\begin{align}
    \label{eq:covmultsplitFRthenWexp}
    P_1 &= C_\pi  + \left(e^{\Gamma \gamma}(C_0-C_\pi )^{-1} e^{\Gamma\gamma} + (e^\gamma-1)C_\pi^{-1} e^{2\gamma C_\pi^{-1}}\right)^{-1}\\
\label{eq:meanmultsplitFRthenWexp}
   a_1 &=  m_\pi +  (P_1 - C_\pi ) e^{\gamma C_\pi^{-1}}(C_0 - C_\pi )^{-1}(m_0 - m_\pi).
\end{align}
for the mean $a_1$ and covariance $P_1$ of a single step of FR-W of size $\gamma$. 

The structure of the mean for both splitting schemes is exactly the same as that obtained for the exact WFR flow with $C_\gamma$ replaced by $Q_1, P_1$ respectively.  This demonstrates that in the Gaussian case the speed-up due to splitting is driven primarily through the improvement in covariance estimation.  In particular, notice the difference in structure between $C_\gamma$ vs $Q_1$ and $P_1$.  The difference due to operator ordering is more subtle, as $Q_1$ and $P_1$ differ only by a single factor $e^{2\gamma C_\pi^{-1}}$.  As will be shown later in this section, this factor does not necessarily lead to acceleration in all cases, and depends on the relationship between $\mu_0$ and $\pi$.

In order to compare the continuous evolution given by~\eqref{eq:WFRGaussexactCt}-\eqref{eq:WFRGaussexactmt} with that given by the discrete time splitting schemes we consider $m_{n\gamma}, C_{n\gamma} $, the mean and covariance respectively at time $t = n \gamma$ under the exact WFR flow. We will thus compare $m_{n\gamma}, C_{n\gamma} $ with the mean and variance of the W-FR (and FR-W) split at step $n$ with step size $\gamma$.

\begin{figure}[htbp!]
	\centering
	\begin{tikzpicture}[every node/.append style={font=\normalsize}]
 \node (img1) {\includegraphics[width = 0.45\textwidth]{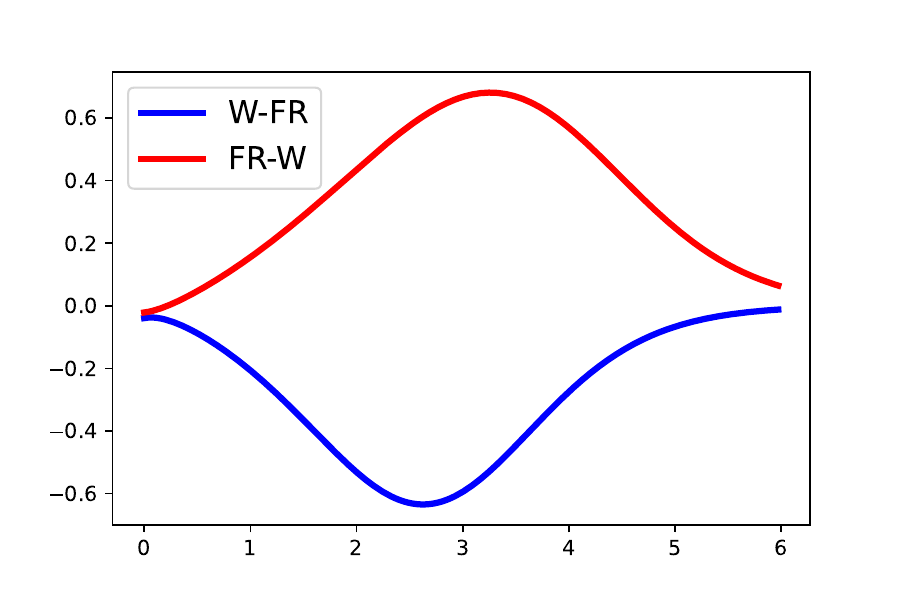}};
   \node[below=of img1, node distance = 0, anchor = center, yshift = 1cm] {$\gamma$};
   \node[left=of img1, node distance = 0, rotate=90, anchor = center, yshift = -0.8cm] {$\textrm{Split}\,\KL-\KL(\mu_\gamma||\pi)$};
  \node[right=of img1, node distance = 0, xshift = -1cm] (img2) {\includegraphics[width = 0.45\textwidth]{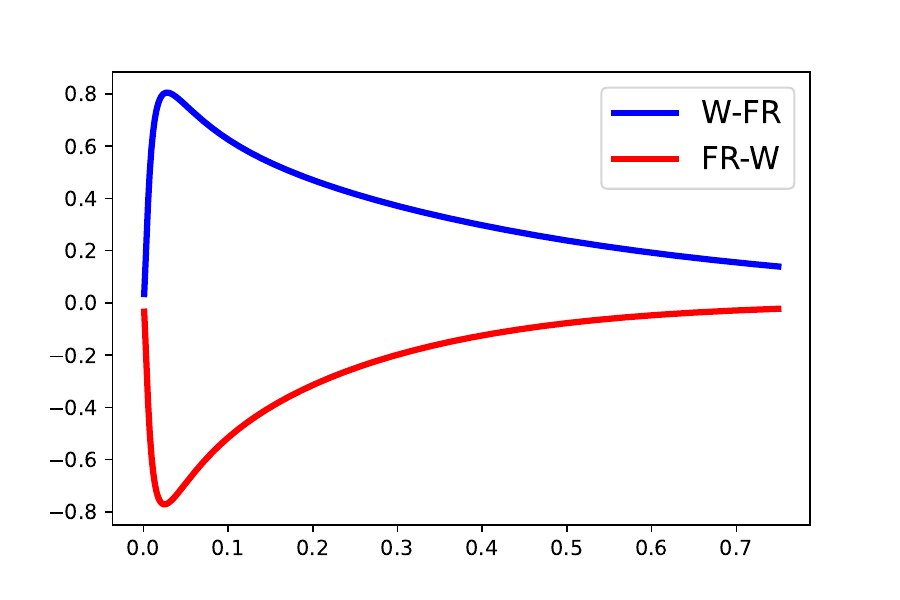}};
\node[below=of img2, node distance = 0, anchor = center, yshift = 1cm] {$\gamma$};
\node[left=of img2, node distance = 0, rotate=90, anchor = center, yshift = -0.8cm] {$\textrm{Split}\,\KL-\KL(\mu_\gamma||\pi)$};
\node[above=of img1, node distance = 0, yshift = -1.2cm] {$m_\pi = 20, C_\pi  = 100, m_0 = 0, C_0 = 1$};
\node[above=of img2, node distance = 0, yshift = -1.2cm] {$m_\pi = 20, C_\pi  = 1, m_0 = 0, C_0 = 100$};
	\end{tikzpicture}
\caption{Difference in $\KL$ for a single time step $\gamma$ for W-FR split and FR-W on 1D Gaussians. Left: Target more diffuse than initial distribution ($m_\pi = 20, C_\pi  = 100, m_0 = 0, C_0 = 1$). Right: Target more concentrated than initial distribution ($m_\pi = 20, C_\pi  = 1, m_0 = 0, C_0 = 100$).}
\label{fig:Gauss1DKLdiffgammaset1}
\end{figure}

Before presenting the main analytic results, consider Figure \ref{fig:Gauss1DKLdiffgammaset1} which compares KL distance from $\pi$ after a single sequential split step and the same for the exact WFR in the univariate Gaussian setting.  More specifically, it shows 
$\KL(\nu_1||\pi) - \KL(\mu_\gamma||\pi)$ (blue line) and $\KL(\eta_1||\pi) - \KL(\mu_\gamma||\pi)$ (red line) for a single sequential split step as a function of step size $\gamma$.  
The KL is calculated exactly using the formula for Gaussians \eqref{eq:kl_gaussian} and mean and covariance using the equations above.  We consider the case where the target is more diffuse than the initial distribution (left plot, $C_\pi  = 100, m_\pi = 20, C_0 = 1, m_0 = 0$) and the opposite case (right plot, $C_\pi  = 1, m_\pi = 20, C_0 = 100, m_0 = 0$).  Figure \ref{fig:Gauss1DKLdiffgammaset1} shows that the improvement in KL due to splitting for larger $\gamma$ depends on the both the order of the operators \textit{and} the on the relationship between the target and initial distribution.  Specifically, in the univariate Gaussian case, when the target variance is \textit{larger} than the initial variance, the ordering W-FR leads to a speed-up over the exact dynamics for all $\gamma > 0$, see Figure \ref{fig:Gauss1DKLdiffgammaset1} left panel.  As expected for $\gamma$ large enough, the difference approaches zero as both the W and FR flows individually have been evaluated for a sufficiently long time so that the resulting densities are close to $\pi$.  Conversely, when the target variance is \textit{smaller} than the initial variance the ordering FR-W is superior, see Figure \ref{fig:Gauss1DKLdiffgammaset1} right panel.

These numerical results are somewhat intuitive; if $\pi$ is more diffuse than $\mu_0$, applying the W operator first allows to increase variance and transform the initial distribution into a more diffuse one; if instead $\pi$ is more concentrated than $\mu_0$ applying the FR operator first allows to shrink the variance and returns a more concentrated distribution since the ratio $\pi/\mu_0$ will be small over most of the support.  We will make this intuition more precise in the remainder of this section and extend to the multivariate setting, where we show that the difference in initial and target means also plays a role.

\begin{figure}[htbp!]
	\centering
	\begin{tikzpicture}[every node/.append style={font=\normalsize}]
 \node (img1) {\includegraphics[width = 0.4\textwidth]{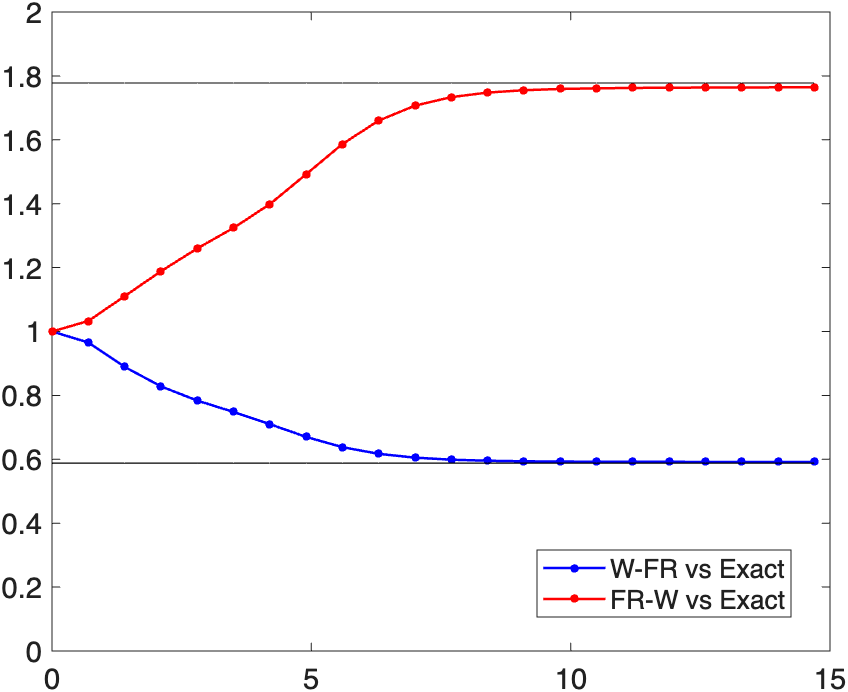}};
   \node[below=of img1, node distance = 0, anchor = center, yshift = 1cm] {$ t $};
   \node[left=of img1, node distance = 0, rotate=90, anchor = center, yshift = -0.8cm] {$\frac{\textrm{Split}\,\KL}{\textrm{Exact}\,\KL}$};
  \node[right=of img1, node distance = 0, xshift = -0.4cm] (img2) {\includegraphics[width = 0.48\textwidth]{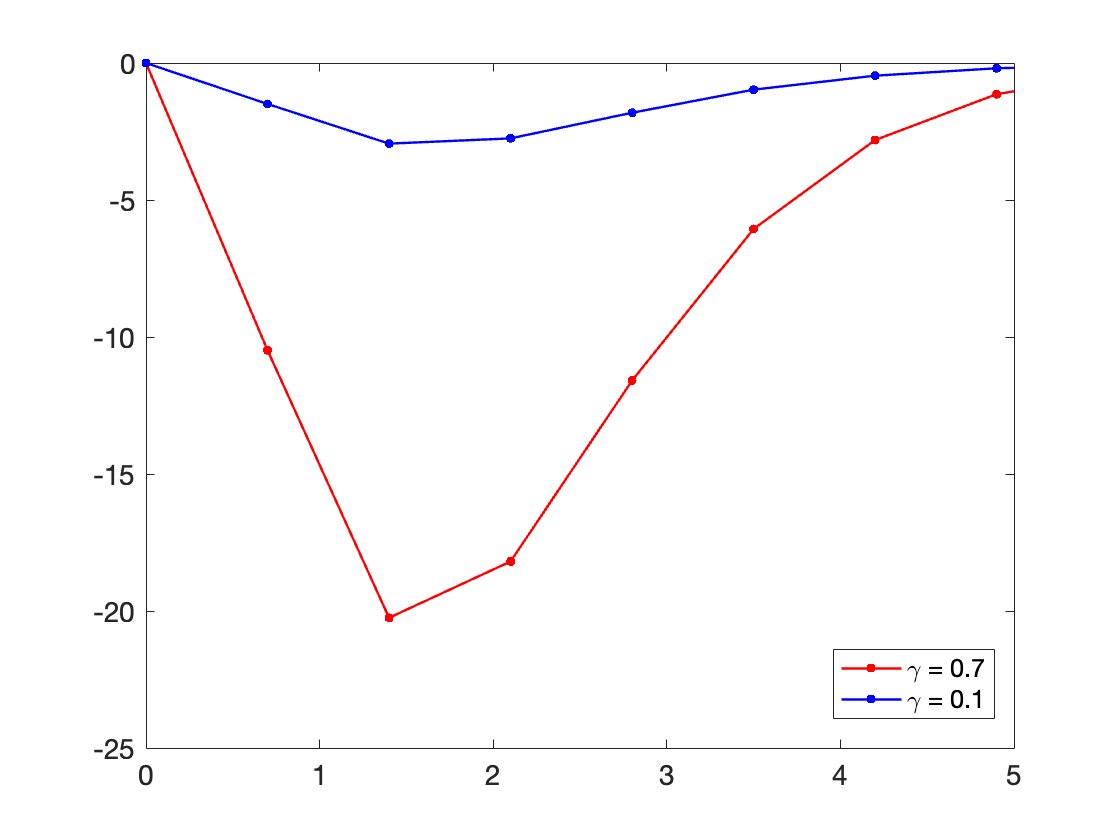}};
\node[below=of img2, node distance = 0, anchor = center, yshift = 1cm] {$ t $};
\node[left=of img2, node distance = 0, rotate=90, anchor = center, yshift = -0.8cm] {$\textrm{W-FR} \, \KL - \textrm{Exact} \, \KL$};
	\end{tikzpicture}
\caption{
Left: KL from $n$-step W-FR (FR-W) scheme divide by KL from exact WFR as a function of $t = n \times \gamma$, $\pi$ is a 10D Gaussian and $\gamma = 0.7$.  The horizontal black line corresponds to \eqref{eq:KLsplitratio}. 
Right: Difference between KL from $n$-step W-FR scheme and KL from exact WFR as a function of $t$ for $\gamma = 0.1$ (blue) and $\gamma = 0.7$ (red). The case $\gamma= 0.1$ corresponds to $n$ steps of alternating \emph{one} W step and \emph{one} FR step. The case $\gamma = 0.7$ corresponds to \emph{7 consecutive steps} of W flow evaluated over time horizon $0.1$ followed by \emph{7 steps} of FR flow over time $0.1$  In both cases, the total number of steps (of size $0.1$) is $n =50$.
This highlights the significant improvement in KL simply by modifying the ordering of the steps without changing the total number of operator evaluations (i.e. same cost).
}
\label{fig:Gauss10D}
\end{figure}

As a first step, we present an equality for the continuous time KL decay for the WFR PDE \eqref{eq:WFRpde}, which to the best of our knowledge is missing from the literature.  
The exact covariance and mean evolution in \eqref{eq:bgammaWthenFR}-\eqref{eq:QgammaWthenFR} for the W-FR split and \eqref{eq:meanmultsplitFRthenWexp}-\eqref{eq:covmultsplitFRthenWexp} for the FR-W split can be used in a recursive fashion to characterise the decay of $\KL$ distance to $\pi$ of the split schemes compared to the exact WFR flow.  We will make use of the following quantities throughout, 
\begin{align}
    \label{eq:defncovdiff}
    E_n := C_{n\gamma} - C_\pi ; \quad &E_n^\beta := Q_{n} - C_\pi;   \quad E_n^\alpha := P_{n} - C_\pi;   \\
    \label{eq:defnmeandiff}
    \varepsilon_n := m_{n\gamma} - m_\pi; \quad & \varepsilon_n^\beta := b_{n} - m_\pi; \quad \varepsilon_n^\alpha := a_{n} - m_\pi.
\end{align}
Consider Figure \ref{fig:Gauss10D} (left) which shows the ratio of split $\KL$ to exact WFR $\KL$ for a 10D Gaussian target  with $\gamma = 0.7$.  For this particular case, the W-FR scheme offers a speed-up over the exact dynamics, where $\KL$ from the split scheme is $\approx 60\%$ of $\KL$ in the exact case for large $t$.  Conversely, a FR-W scheme leads to a reduction in convergence speed (see Figure \ref{fig:Gauss10D} (right)).  Importantly, while the ratio stabilises at approximately $1.8$, both $\KL(\eta_n||\pi)$ and $\KL(\mu_{n\gamma}|| \pi)$ approach zero so that there is no time asymptotic bias.  As emphasised in Section \ref{sec:intro},   Figure \ref{fig:Gauss10D}(right) demonstrates the benefit of modifying the ordering of a sequence of operations, without changing the total number of evaluations. 

In the remainder of this section, we characterise KL decay and the properties giving rise to acceleration/deceleration.  
The following result characterises KL decay as a function of $n$ for the exact and sequential split schemes in the multivariate Gaussian case.  The proof can be found in Appendix \ref{sec:proofKL}. 

\begin{lemma}
    \label{lem:KLexactWFR}
    \textbf{KL decay, multivariate Gaussian.} Consider the multivariate Gaussian setting, i.e. $\mu_0(x) = \mathcal{N}(x; m_0, C_0)$ and $\pi(x) = \mathcal{N}(x; m_\pi, C_\pi )$ with $C_0, C_\pi$ strictly positive definite and $E_0:= C_0 - C_\pi$ invertible.   
    Assume the exact WFR \eqref{eq:WFRpde}, W-FR split scheme \eqref{eq:sequential_split} and FR-W split scheme \eqref{eq:sequential_splitFRthenW} are initialised with the same $C_0, m_0$ (i.e. $E_0, \varepsilon_0$ fixed across methods), so that $\mu_0 = \nu_0 = \eta_0$.  Recall $\Gamma := C_\pi^{-1} + \frac{1}{2} I$. Define also  
    \begin{align*}
        \phi_n(A) &:= \frac{1}{2} \left[ -\log \det(I + C_\pi^{-1} A) +  Tr[C_\pi^{-1}A] + \varepsilon_0^\top E_0^{-1} e^{n \gamma C_\pi^{-1}} A C_\pi^{-1} A e^{n \gamma C_\pi^{-1}} E_0^{-1} \varepsilon_0  \right], \\
        J_n(B) &:= e^{-n \gamma \Gamma} \left(E_0^{-1}  + B\, C_\pi^{-1}( I - e^{-2n \gamma \Gamma}) \right)^{-1}e^{-n \gamma \Gamma},
    \end{align*}
    where $A$ denotes any $d \times d$ symmetric matrix satisfying $C_\pi+A\succ0$ and $B \in \mathbb{R}^{d \times d}$ is such that the inverse in $J_n(B)$ exists. 
    Finally, let 
    \begin{align}
     \label{eq:defnomega}
        \Omega &:= \frac{1}{2} \Gamma^{-1} \\
         \label{eq:defnomegabeta}
        \Omega^\beta &:= (1-e^\gamma)(I - e^{2 \gamma \Gamma})^{-1}, \\
          \label{eq:defnomegaalpha}
        \Omega^\alpha &:= (1-e^\gamma)(I - e^{2 \gamma \Gamma})^{-1}e^{2\gamma C_\pi^{-1}} = (1 - e^\gamma) (e^{-2\gamma C_\pi^{-1}} - e^{\gamma })^{-1}.
    \end{align}
    The solution of the exact WFR PDE \eqref{eq:WFRpde} at time $t = n\gamma$ satisfies
    \begin{align}
        \nonumber
        \KL(\mu_{n \gamma}||\pi) = \phi_n(E_n), &\enskip E_n =   J_n(\Omega) 
    \end{align}
    Similarly, for the split schemes, $\KL(\nu_n||\pi)  = \phi_n(E_n^\beta), \enskip E_n^\beta =   J_n(\Omega^\beta),$ and $\KL(\eta_n||\pi)  = \phi_n(E_n^\alpha), \enskip E_n^\alpha =   J_n(\Omega^\alpha)$. 
   
\end{lemma}

Despite the considerable difference in structure of $C_\gamma$ vs $Q_1$ or  $P_1$  (c.f. \eqref{eq:WFRGaussexactCt} with $t=\gamma$ and \eqref{eq:QgammaWthenFR}, \eqref{eq:covmultsplitFRthenWexp}), the $n$-step iterations lead to expressions that more clearly highlight similarities between the exact and sequential splitting.  More specifically, the covariance difference quantities $E_n$ and $E_n^\beta, E_n^\alpha$ are identical up to factors depending on $\Omega$ and $\Omega^\beta, \Omega^\alpha$ (see \eqref{eq:Enbeta} - \eqref{eq:Enexpr}). Lemma \ref{lem:KLexactWFR} shows that differences in KL decay between the exact and split flows are controlled entirely by these multiplicative factors.  

This brings us to our main result of this section, the proof of which can be found in Appendix \ref{sec:proofKLsplitvsexact}.  It quantifies acceleration or deceleration over the exact WFR dynamics due to splitting in terms of KL decay.

\begin{proposition}
\label{prop:KLcomp}
    \textbf{Convergence speed, multivariate Gaussian.} Assume the same conditions as in Lemma \ref{lem:KLexactWFR} and let $||\cdot||_F$ denote the Frobenius norm. 
    Then the following holds:  
    \begin{enumerate}[label=(\roman*)]
    \item Suppose $E_0 \succ 0$.  Then $E_n \succ 0, E_n^\beta \succ 0$ and $E_n^\alpha \succ 0$ for all $n=1,2,3 \dots$.  Furthermore, $||E_n^\alpha||_F < ||E_n||_F < ||E_n^\beta||_F$ for all $n = 1, 2, 3, \dots$.
    \item Suppose $E_0 \prec 0$.  Then $E_n \prec 0, E_n^\beta \prec 0$ and $E_n^\alpha \prec 0$ for all $n=1,2,3 \dots$.  Furthermore, $||E_n^\alpha||_F > ||E_n||_F > ||E_n^\beta||_F$ for all $n = 1, 2, 3, \dots$.
     \item Assume the conditions of (i) or (ii) and that $m_0 \neq m_\pi$.  Let $p_1$ denote the eigenvector corresponding to the smallest eigenvalue of $C_\pi^{-1}$ and suppose $p_1^\top (I  + E_0\Omega^\beta C_\pi^{-1})^{-1} \varepsilon_0 \neq 0$.  Then the asymptotic (large $n$) ratio of the $\KL$ distance to $\pi$ of the W-FR split scheme solution with respect the solution of the exact WFR is given by 
     \begin{align}
     \label{eq:KLsplitratio}
      \lim_{n \rightarrow \infty} \frac{\KL(\nu_n||\pi)}{\KL(\mu_{n \gamma}||\pi)} = \frac{p_1^\top (I  + E_0\Omega^\beta C_\pi^{-1})^{-1} \varepsilon_0 \varepsilon_0^\top (I + E_0\Omega^\beta C_\pi^{-1})^{-\top} p_1}{p_1^\top (I  + E_0\Omega C_\pi^{-1})^{-1} \varepsilon_0 \varepsilon_0^\top (I  + E_0\Omega C_\pi^{-1})^{-\top} p_1},
\end{align}
where $\Omega$ and $\Omega^\beta$ are as defined in \eqref{eq:defnomega} and \eqref{eq:defnomegabeta} respectively.
Likewise, for the FR-W split scheme, the same limiting ratio for $\frac{\KL(\eta_{n}||\pi)}{\KL(\mu_{n \gamma}||\pi)}$ holds with $\Omega^\beta$ replaced by $\Omega^\alpha$ as defined in \eqref{eq:defnomegaalpha}.   
     \end{enumerate}   
\end{proposition}

Notice that in dimension 1, the limiting ratio in \eqref{eq:KLsplitratio} is independent of $\varepsilon_0$, and the speed-up depends only on the initial covariance relative to the target covariance.  In particular,  if $C_0>C_\pi$ then FR-W will converge faster than the exact WFR dynamics, conversely, if $C_\pi>C_0$ then W-FR will be faster.  In the multivariate case however, the mean difference $\varepsilon_0$ plays an important role also and the sign of $E_0$ alone cannot be used to determine where the ratio in \eqref{eq:KLsplitratio} is strictly larger or smaller than 1.  Instead, for a chosen step size $\gamma$, if 
\begin{align*}
    |p_1^\top (I + E_0 \Omega^\beta C_\pi^{-1})^{-1} \varepsilon_0| <|p_1^\top (I + E_0 \Omega C_\pi^{-1})^{-1} \varepsilon_0|
\end{align*}
then W-FR will converge faster than the exact dynamics (similarly for FR-W with $\Omega^\beta$ replaced by $\Omega^\alpha$).

\section{Preservation of log-concavity and Convergence speed of W-FR}

As a first step towards generalising the results of Section \ref{sec:gaussian}, we consider the case of log-concave target distributions. 
Existing $\KL$ decay results for the W flow and FR flow developed in the literature \cite{chen2023sampling,Lu2019, domingo-enrich2023an, lu2023birth, carrillo_fisher-rao_2024} cannot be straightforwardly utilised to characterise convergence of the split schemes.  In particular, these exponential decay bounds for the W part and the FR part would not allow us to isolate changes to convergence speed depending on the order as the exponents simply add.  Hence, the first variation results developed in Proposition~\ref{lem:seqsplitpdeWthenFR} and~\ref{lem:seqsplitpdeFRthenW} will be critical in characterising how convergence speed is affected due to splitting, and the ordering of operators. Due to the intractability of \eqref{eq:seqsplitpdeFRthenW}, we primarily focus on the W-FR scheme as a first step towards understanding the more general log-concave case.

\subsection{Preservation of log-concavity}
\label{sec:logconcave}

We first provide a result demonstrating that log-concavity is preserved under the evolution of the W-FR and FR-W split schemes. 
To do so, we utilise recent results that quantify the time horizon over which log-concavity is preserved under the W flow \cite[Lemma 1]{us_bernoulli}.
We first state the necessary assumptions:
\begin{assumption}
\label{ass:logconcave} In addition to Assumption~\ref{ass:lsi}, the following conditions hold  
\begin{enumerate}[label=(\alph*)]
\item $\pi(x) \propto e^{-V_\pi(x)}$ is $\alpha_\pi$-strongly log-concave and $L_\pi$-smooth, i.e. there exists an $L_\pi\geq \alpha_\pi > 0$ such that $ L_\pi I \succeq \nabla^2 V_\pi(x) \succeq \alpha_\pi I$ for all $x \in \mathbb{R}^d$;
\item $\mu_0(x) \propto e^{-V_0(x)}$ is $\alpha_0$-strongly log-concave and $L_0$-smooth, i.e. there exists an $L_0\geq \alpha_0 > 0$ such that $ L_0 I \succeq \nabla^2 V_0(x) \succeq \alpha_0 I$ for all $x \in \mathbb{R}^d$.
\end{enumerate}
\end{assumption}
This assumption requires Lipschitz continuity of the gradients of both the target and initial potential.  While strong, this assumption is classical in the study of the W flow.
The assumption on $\mu_0$ is mild since $\mu_0$ is often user-chosen and guarantees that $\nabla V_0$ is Lipschitz continuous with constant $L_0$. 

\begin{assumption}
\label{ass:WLC} Given Assumption~\ref{ass:logconcave}, the following conditions hold.  
\begin{enumerate}[label=(\alph*)]
\item \label{ass:WLC1} $V_0 - \frac{(1 + \delta)}{2}V_\pi$ is strongly convex with parameter $\alpha_d > 0$ for some specified $0 < \delta < 1$.   
\item Denote by $R:= -\frac{1}{2}\Delta V_\pi + \frac{1}{4}|\nabla V_\pi|^2$. Also define $\mathcal{H}:= R + V_\pi$.  Assume $\mathcal{H}$  is strongly convex with parameter $\alpha_h > 0$. 
\end{enumerate}
\end{assumption}

The first condition requires the initial density to be ``sufficiently'' strongly log-concave w.r.t. the target density. As $\mu_0$ is often user chosen, this assumption is not overly restrictive.  The second condition is a uniform coercivity condition on the so-called 
Schroedinger potential, which ensures the target potential is strongly confining.  It implies $\nabla^2 R = \nabla^2\mathcal{H}-\nabla^2V_\pi \succeq (\alpha_h-L_\pi)I$.
Thus, when $\alpha_h\geq L_\pi$, the W-flow estimate does not lose curvature through $R$; when $\alpha_h<L_\pi$, only a finite W-flow horizon is guaranteed and the splitting step size must be restricted accordingly.

Under these assumptions, similar techniques to those used in \cite{us_bernoulli} can be used to obtain a uniform in $n$ preservation of log-concavity for the W-FR split scheme, so long as $\gamma$ is not too large, as shown in the following proposition.  The proof is essentially identical to that in  \cite[Appendix A.1]{us_bernoulli}, up until the limit $\tau_i \rightarrow 0$ is taken. An equivalent result can be obtained for the FR-W split.

\begin{proposition}
    \label{corr:logconcsplit}
    Let Assumption~\ref{ass:WLC} hold. Then $\nu_{n}$, the solution of the W-FR split scheme \eqref{eq:sequential_split} with step size $\gamma$, is $\beta_n$-strongly log-concave uniformly in $n$ whenever $\alpha_h - L_\pi > 0$.  When $\alpha_h - L_\pi < 0$, the same is true, as long as $b:= \sqrt{\frac{|\alpha_h - L_\pi|}{2}}$ satisfies  
    \begin{align}
        \label{ass:b2}
        b^2 < \tfrac{\alpha_\pi}{4} 
    \end{align} 
    and
        $0< \gamma < \min(\tau_1^\ast, \tau_\infty^\ast) $
    where $\tau_1^\ast$ is given by
    \begin{align*}
    \tau^\ast_i =  \frac{1}{2b} \tan^{-1} \left(  \frac{(1 - e^{-\tau_{i-1}^\ast})\alpha_\pi}{2b}  \right), \enskip i = 1, 2, 3, \dots  
    \end{align*}
    and $\tau^\ast_\infty$ is the nontrivial solution of the fixed point equation $\tau^\ast_\infty = \frac{1}{2b} \tan^{-1} \left(  \frac{(1 - e^{-\tau^\ast_\infty})\alpha_\pi}{2b}  \right)$.
Furthermore, $ \beta_n > \frac{\alpha_\pi}{2}, \quad n = 1, 2, 3, \dots$.
\end{proposition}
The condition \ref{ass:b2} requires that the loss of strong log-concavity due to the W flow, quantified by $b^2$ is sufficiently weak relative to the target’s strong log-concavity $\alpha_\pi$. It ensures that the Fisher–Rao contribution is strong enough to compensate for this loss, yielding uniform-in-time preservation of strong log-concavity under the WFR flow.  Such a condition easily holds, for example, in 1D Gaussian targets where it translates to $C_\pi > 1$.

\subsection{Convergence speed of W-FR}   
\label{sec:convlogconc}
Decay rates in KL for the exact WFR have been studied in \cite{Lu2019, domingo-enrich2023an, lu2023birth} and combine the rate of decay of the W flow (e.g. \cite{chewi2024analysis}) with that of the FR flow \cite{chen2023sampling}.  The symmetrised KL, defined by $J(\mu, \pi):= \KL(\mu||\pi) + \KL(\pi||\mu)$
for any $\mu \ll \pi$ and also $\pi \ll \mu$, is a more natural object to study convergence to $\pi$, particularly under the Fisher-Rao geometry. This is due to the fact that the reverse KL, $\KL(\mu||\pi)$, is not geodesically convex under the Fisher-Rao geometry \cite{carrillo_fisher-rao_2024}.  It has recently been shown \cite{us_bernoulli} using functional inequalities from \cite{carrillo_fisher-rao_2024} that under Assumptions \ref{ass:logconcave} and \ref{ass:WLC},
\begin{align}
    \label{eq:WFRsymrate}
    J(\mu_{n \gamma}, \pi) < J(\mu_0, \pi) e^{-n\gamma(\alpha_\pi +1 )}.
\end{align}
In this section, we study the convergence rate of the W-FR split scheme using the variational formula \eqref{eq:seqsplitpdeWthenFR}.
From Proposition \ref{lem:seqsplitpdeWthenFR}, it is clear that the behaviour of $J(\nu_{n}, \pi)$ for the W-FR split scheme will depend on the perturbation term $f_{\text{P}}$.  As shown below, the contribution of this term will depend on the sign of $\text{Cov}_{\tilde{\nu}_\tau}\left(g(\nu_1), \left|\nabla g(\nu_1)\right|^2\right) $ where $\tilde{\nu}_\tau \propto  \nu_1^{(1-\tau)} \pi^\tau$ for all $\tau \in [0,1]$. 
This quantity measures the covariation between the log density ratio (when it is pointwise positive, it means $\nu_1$ is larger than the target density at that point) and the magnitude of the difference in score functions, which is large in regions where the ratio $\frac{\nu_1(x)}{\pi(x)}$ varies rapidly. When this covariance term is negative, regions where $\pi$ dominates $\nu_1$ tend to correspond to regions where local gradients of $\pi$ and $\nu_1$ differ, when measured against the interpolated density $\tilde{\nu}_\tau$. We rely on the following assumption.
\begin{assumption}
    \label{ass:cov0}
    Suppose $\mu_0$ and $\pi$ are chosen such that $\nu_1$, the solution of the W-FR split scheme after a single step of size $\gamma$ satisfies  
    \begin{align}
         \label{eq:covariance_assumption}\text{Cov}_{\tilde{\nu}_\tau}\left(g(\nu), \left|\nabla g(\nu)\right|^2\right) < 0,
    \end{align}
    where $g$ is as defined in \eqref{eq:geqn} and $\tilde{\nu}_\tau \propto  \nu^{1-\tau} \pi^\tau$ for all $\tau \in [0,1]$. 
\end{assumption}

Deriving conditions on $\mu_0$ and $\pi$ such that Assumption \ref{ass:cov0} (and related inequalities) hold is an active area of research \cite{egozcue2011covariance, bonnefont2024covariance, egozcue2009some}. Inequalities of the form $\text{Cov}_{\nu}(\alpha(X), \beta(X))<0$ for general measures $\nu$ and functions $\alpha, \beta$ can be obtained under strong co-monotonicity assumptions on $\alpha, \beta$ \cite{armstrong1993chebyshev}.  In Appendix \ref{app:covariance}, we demonstrate it holds for $\nu(x) = \mathcal{N}(x; b, Q)$ and $\pi(x) = \mathcal{N}(x; m_\pi, C_\pi)$ with $\textrm{sym}(C_\pi Q^{-1}) \succ I$.  
We speculate that similar results should hold when $g$ is strongly concave in $x$. A first step in this direction is examined in the 1D symmetric case in Lemma~\ref{app:cov1D}.   Establishing similar results for more relaxed conditions is left for future work. 

\begin{figure}[htbp!]
	\centering
	\begin{tikzpicture}[every node/.append style={font=\normalsize}]
 
  \node (img2) {\includegraphics[width = 0.45\textwidth]{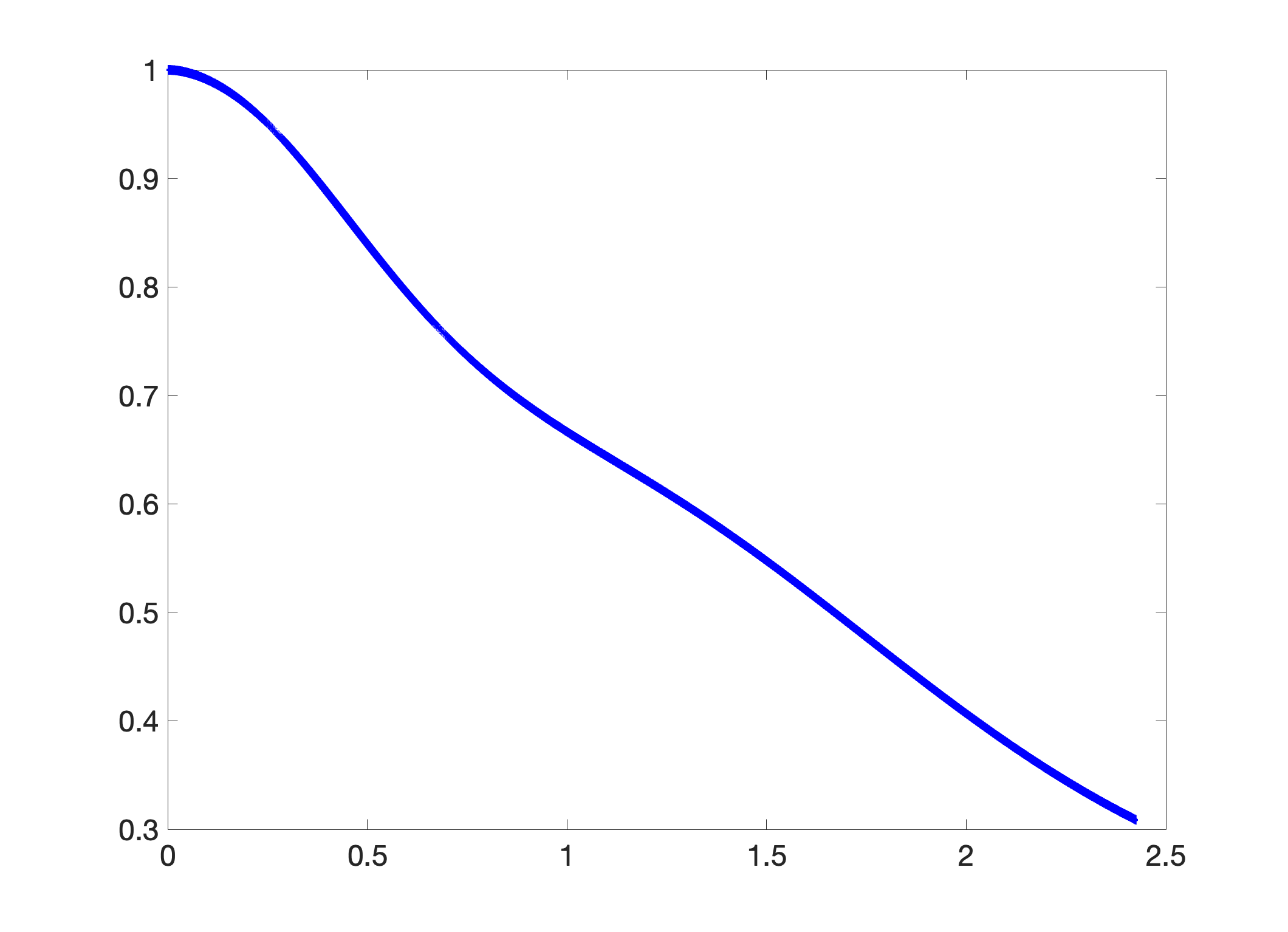}};
\node[below=of img2, node distance = 0, anchor = center, yshift = 1cm] {$  \gamma$};
\node[left=of img2, node distance = 0, rotate=90, anchor = center, yshift = -0.8cm] {$\frac{\textrm{Split}\,\KL(\nu_{1}||\pi)}{\KL(\mu_{\gamma}||\pi)}$};
\node[above=of img2, node distance = 0, yshift = -1.2cm] {1D Gaussian mixture (2 component)};
	\end{tikzpicture}
\caption{Ratio of KL from \textit{one} step of W-FR scheme to KL from continuous time WFR as a function of step size $\gamma$ where $\pi$ is a 2 component univariate Gaussian mixture given by $\pi(x) =  0.5 \, \mathcal{N}(x; -3, 4) + 0.5 \,\mathcal{N}(x; 3, 1)$.  Initial density is $\mu_0(x) = \mathcal{N}(x; 0, 0.2)$.  }
\label{fig:Gauss10Dandmixture}
\end{figure}

The below proposition establishes a decay property for the split W-FR flow for log-concave $\mu_0, \pi$, yielding a rate which is faster than the exact WFR rate (c.f. \eqref{eq:WFRsymrate}).  As this result only pertains to an upper bound on the symmetrised KL, the second result \eqref{eq:splitfast} shows that after a single step, the result of the W-FR split scheme is closer to the target (in terms of symmetrised KL) than the exact WFR flow.  Characterisation of decay for more general targets is a natural next step.  We hypothesise that when the target variance is larger than the initial variance (similar to the Gaussian case with condition $\textrm{sym}(C_\pi Q^{-1}) \succ I$), the W-FR split scheme will yield a speed-up. As motivating evidence of a non-log concave case, consider the 1D Gaussian mixture in Figure~\ref{fig:Gauss10Dandmixture} where speed-up is achieved over a single step with the W-FR split scheme.

\begin{proposition}
\label{prop:symKLsplit}
     Assume the conditions of Proposition~\ref{corr:logconcsplit} and Assumption \ref{ass:cov0}.  Then the following decay result holds for $\nu_{n }$ the solution of the W-FR split scheme for $1\leq n\leq n_m$
    \begin{align}
    \label{eq:convratesplit}
        J(\nu_{n }, \pi) < J(\mu_0, \pi) e^{-n\gamma (\alpha_\pi +1) - c(n_m)(e^\gamma - \gamma - 1) }
    \end{align}
    for some $c > 0$ dependent on $\gamma$ and $n_m$.  Furthermore, there exists $\gamma^\ast > 0$ such that for all $\gamma \in (0, \gamma^\ast)$
    \begin{align}
        \label{eq:splitfast}
        J(\nu_1, \pi) - J(\mu_\gamma, \pi) < 0.
    \end{align} 
\end{proposition}
\begin{proof}
    For the first claim, \eqref{eq:convratesplit}, consider the evolution along the W-FR PDE~\eqref{eq:seqsplitpdeWthenFR}.
Given the assumptions, we may interchange differentiation and integration to obtain 
\begin{align}
    \label{eq:J1diff}
    \frac{d}{d\gamma} J(\nu_1, \pi) & = \int \left( \log \frac{\nu_1}{\pi} -\frac{\pi}{\nu_1} \right)(x) \left(f_\W(\nu_1(x)) + f_\F(\nu_1(x)) + (e^\gamma -1)f_{\text{P}}(\nu_1(x)) \right)  dx \\
    \nonumber 
    &= -(\mathcal{I}(\nu_1|| \pi) + \mathcal{I}(\pi||\nu_1)) - J(\nu_1, \pi)  -\text{Var}_{\nu_1} \left( \log \frac{\nu_1}{\pi} \right)  \\
    \nonumber 
    &+ (e^\gamma -1) \left[\text{Cov}_{\nu_1}\left(\log \frac{\nu_1}{\pi}, \left|\nabla\log \frac{\nu_1}{\pi}\right|^2\right) -\mathcal{I}(\pi||\nu_1) + \mathcal{I}(\nu_1||\pi) \right] \\
    \nonumber 
    &\leq -(\mathcal{I}(\nu_1|| \pi) + \mathcal{I}(\pi||\nu_1)) - J(\nu_1, \pi) \\
    \nonumber 
    &+(e^\gamma-1)\left[\text{Cov}_{\nu_1}\left(\log \frac{\nu_1}{\pi}, \left|\nabla\log \frac{\nu_1}{\pi}\right|^2\right) -\mathcal{I}(\pi||\nu_1) + \mathcal{I}(\nu_1||\pi) \right].
\end{align}
The term $\text{Cov}_{\nu_1}\left(\log \frac{\nu_1}{\pi}, \left|\nabla\log \frac{\nu_1}{\pi}\right|^2\right)  < 0$ due to Assumption \ref{ass:cov0} with $\tau = 0$.  We will now show that $-\mathcal{I}(\pi||\nu_1) + \mathcal{I}(\nu_1||\pi) < 0$ also under Assumption \ref{ass:cov0}. Recall $\nu_1 \propto e^{-V(x)}$ with $V(x)$ strongly convex due to Corollary \ref{corr:logconcsplit}. Define $U_\tau(x) := (1-\tau)V(x) + \tau V_\pi(x)$ and  $\tilde{\nu}_\tau = \frac{e^{-U_\tau(x)}}{\int e^{-U_\tau(y)}dy}$. Clearly $\tilde{\nu}_0 = \nu_1$ and $\tilde{\nu}_1 = \pi$. Also define $F_\tau := \int |\nabla g(\nu_1(x))|^2 \tilde{\nu}_\tau(x)dx$ where $g(\nu_1) = \log \frac{\nu_1}{\pi}$ and once again we have $F_1 = \mathcal{I}(\pi||\nu_1)$ and $F_0 = \mathcal{I}(\nu_1||\pi)$. 
Note that since  $\partial_\tau U_\tau = -V + V_\pi = g(\nu_1)$, 
\begin{align*}
    \partial_\tau \tilde{\nu}_\tau 
    & = -\partial_\tau U_\tau \tilde{\nu}_\tau + \tilde{\nu}_\tau  \int \partial_\tau U_\tau (x) \tilde{\nu}_\tau (x)dx  = -g(\nu_1) \tilde{\nu}_\tau + \tilde{\nu}_\tau \mathbb{E}_{\tilde{\nu}_\tau} \left[  g(\nu_1) \right] 
\end{align*}
Then differentiating $F_\tau$ and swapping order of derivative and integral (permitted under the assumptions) yields  
\begin{align*}
    \frac{d F_\tau}{d\tau} & = \int |\nabla g(\nu_1(x))|^2 \partial_\tau \tilde{\nu}_\tau(x)dx \\
    &= \int |\nabla g(\nu_1(x))|^2 \left(-g(\nu_1) \tilde{\nu}_\tau + \tilde{\nu}_\tau  \mathbb{E}_{\tilde{\nu}_\tau} \left[ g(\nu_1) \right] \right)dx\\
    &= -\text{Cov}_{\tilde{\nu}_\tau} \left( |\nabla g(\nu_1)|^2, g(\nu_1)\right) .
\end{align*}
Then under Assumption \ref{ass:cov0}, we have that $\frac{d F_\tau}{d\tau} > 0$ for all $\tau \in [0,1]$, then $F_0 < F_1$ so that we immediately obtain $-\mathcal{I}(\pi||\nu_1) + \mathcal{I}(\nu_1||\pi) < 0$.  Inserting this back into the differential inequality yields 
\begin{align*}
     \frac{d}{d\gamma} J(\nu_1, \pi) &<  -(\mathcal{I}(\nu_1|| \pi) + \mathcal{I}(\pi||\nu_1)) - J(\nu_1, \pi) - (e^\gamma -1)  \sigma_1(\gamma) 
\end{align*}
for some $\sigma_1(\gamma) > 0$ potentially dependent on $\gamma$.
Proceeding as in \cite[Proposition 1]{us_bernoulli} using the log-concavity of $\pi$ and $\nu_1$ established in Proposition~\ref{corr:logconcsplit} we obtain 
\begin{align*}
    J(\nu_1, \pi) &< J(\mu_0, \pi) e^{-\gamma (\alpha_\pi +1) - \int_0^\gamma (e^s -1) \sigma_1(s)ds } 
\end{align*}
where the integral term is strictly negative.  Applying an induction argument with Assumption~\ref{ass:cov0} holding for $\mu_0 = \nu_n$, $n = 2, 3, \dots, n_m$ yields the result with $c = \displaystyle \min_{n = 1, 2, \dots, n_m} \inf_s \sigma_n(s) > 0$. 

For the second claim, \eqref{eq:splitfast}, 
first define $\mathcal{E}_1(\gamma):= J(\nu_1, \pi) - J(\mu_\gamma, \pi)$.  Then  using \eqref{eq:J1diff}, 
\begin{align*}
    \frac{d \mathcal{E}_1}{d\gamma} &= -(\mathcal{P}(\nu_1) - \mathcal{P}(\mu_\gamma)) + (e^\gamma -1) \mathcal{R}(\nu_1) \\
    \mathcal{P}(\rho) &:=  (\mathcal{I}(\rho|| \pi) + \mathcal{I}(\pi||\rho)) + J(\rho, \pi)  + \text{Var}_{\rho} \left( \log \frac{\rho}{\pi} \right)   \\
    \mathcal{R}(\rho) &:= \text{Cov}_{\rho}\left(\log \frac{\rho}{\pi}, \left|\nabla\log \frac{\rho}{\pi}\right|^2\right) -\mathcal{I}(\pi||\rho) + \mathcal{I}(\rho||\pi).
\end{align*}
Our goal is to show that $\frac{d \mathcal{E}_1}{d\gamma}  < 0$ for all $\gamma$ in some time interval to be determined with $\mathcal{E}_1(0) = 0$. 
Then $ \mathcal{E}_1(\gamma) < 0$ for all $\gamma \in (0, \gamma^\ast)$ where the parameter $\gamma^\ast > 0$ is the smallest $\gamma$ such that $\mathcal{E}_1(\gamma^\ast) = 0$ and $\left. \frac{d\mathcal{E}_1}{d\gamma} \right|_{\gamma = \gamma^\ast} \geq 0$.  It holds that
\begin{align*}
    \frac{d^2 \mathcal{E}_1}{d\gamma^2} = -\partial_\gamma \mathcal{P}(\nu_1(\gamma)) + \partial_\gamma \mathcal{P}(\mu_\gamma(\gamma)) + e^\gamma \mathcal{R}(\nu_1(\gamma)) + (e^\gamma - 1)\partial_\gamma \mathcal{R}(\nu_1(\gamma))
\end{align*}
and furthermore that $\left. \frac{d^2 \mathcal{E}_1}{d\gamma^2} \right|_{\gamma = 0} = \mathcal{R}(\nu_1(0))$.  Finally, it holds that $\mathcal{R}(\nu_1(0)) < 0$ since \newline $\text{Cov}_{\nu_1}\left(\log \frac{\nu_1}{\pi}, \left|\nabla\log \frac{\nu_1}{\pi}\right|^2\right)  < 0$ due to Assumption \ref{ass:cov0} with $\tau = 0$ and also $-\mathcal{I}(\nu_1(0)|| \pi) + \mathcal{I}(\pi||\nu_1(0)) < 0$ by the same reasoning as in the proof of the first claim.  Additionally, since $\frac{d^2 \mathcal{E}_1}{d\gamma^2}$ is continuous at $\gamma = 0$, there exists $\gamma^\ast > 0$ such that $\frac{d^2 \mathcal{E}_1}{d\gamma^2} < 0$ for all $0 \leq \gamma < \gamma^\ast$.  Since $\mathcal E_1(0)= \frac{d \mathcal{E}_1(0)}{d\gamma }=0$, it follows that $\mathcal{E}_1(\gamma)<0$ for $0<\gamma<\gamma^\ast$ so that the second claim holds. 

\end{proof}

The analysis of the FR-W split is considerably more involved as the variational formula in Proposition~\ref{lem:seqsplitpdeFRthenW} is less tractable. To control the perturbation terms arising from~\eqref{eq:seqsplitpdeFRthenW} one would in fact need to show that $\text{Cov}_{\tilde{\eta}_\tau} \left( g_k(\eta_1), g(\eta_1)\right)<0$ for all $k\geq 1$,
where $\tilde{\eta}_\tau \propto  \eta_1^{(1-\tau)} \pi^\tau$ for all $\tau \in [0,1]$ and $g_k$ are given in Lemma~\ref{lem:commut}. For $k=1$ this corresponds to $\text{Cov}_{\tilde{\eta}_\tau}\left(g(\eta_1), \left|\nabla g(\eta_1)\right|^2\right) >0$ suggesting that conditions under which the W-FR could lead to a speed-up lead to a decrease in convergence speed for the FR-W split.  This is directly analogous to the Gaussian case (see Proposition~\ref{prop:KLcomp}).

\section{Conclusion \& Future Work}
Operator splitting methods lie at the heart of many gradient flow based algorithms, in particular, for discretisations of the Wasserstein--Fisher--Rao gradient flow.  We examine errors introduced by splitting schemes whereby the Wasserstein and Fisher--Rao operators are solved separately (and exactly).  Such schemes preserve the invariant measure, but produce dynamics that differ from the exact WFR flow.  We develop variational formulae for the split schemes and identify a perturbation which modifies the dynamics.  It is demonstrated that the order in which these operators are evaluated can have a considerable impact on convergence speed, without affecting $\pi$-invariance.  This impact is unsurprisingly more pronounced when the chosen step size is large, but not so large that it is nearly equivalent to evaluating either the Wasserstein or Fisher-Rao flow alone. 
In practical terms, this means that a carefully chosen splitting can exhibit faster convergence to the target than the exact WFR dynamics and thus a wise choice of splitting can reduce time to convergence without increasing the computational cost. 
Our work thus highlights that in the context of gradient flow PDEs for sampling one should focus on numerically approximating the split dynamics rather than the exact flow.

We characterise this behaviour in the Gaussian case, where we analyse the evolution of the $\KL$ to $\pi$ using analytic formulas rather than bounds (Section~\ref{sec:gaussian}). As a first step towards generalisation, we utilise the variational formulae to quantify conditions that can allow for acceleration in the W-FR split case for log-concave targets (see Proposition~\ref{prop:symKLsplit}.

Our analysis assumes that the FR and W operator are evaluated exactly. Due to the large amount of numerical schemes approximating the W flow \cite{durmus2019analysis, roberts_langevin_2002, salim2020wasserstein} and the FR flow \cite{Lu2019, lu2023birth, us, maurais2024sampling}, we leave the investigation of numerical schemes approximating the split dynamics to future work, and in particular, the study of how numerical errors propagate through the split dynamics. 

Finally, we have focused on the impact of operator ordering to improve convergence speed.  Further work would also examine how the split step size $\gamma$ could be chosen to maximise this improvement, as e.g. Figure~\ref{fig:Gauss1DKLdiffgammaset1} demonstrates that the step size should be chosen not too large, but also not too small that the dynamics are close to the exact WFR. 
Furthermore, adaptive strategies for the numerical approximation of the W-FR flow could be borrowed from the sequential Monte Carlo literature on adaptive tempering \cite{jasra2011inference, syed2024optimised} and more recent results on adaptive stepsizes for the W flow \cite{sharrock2025tuning} in investigating optimal step sizes.

\section*{Acknowledgments}


The authors would like to thank Josh Bon, Sam Power, Florian Maire, Daniel Paulin, Upanshu Sharma for helpful discussions and in particular Sinho Chewi for pointing to the reference \cite{Kolesnikov2001} and Andre Wibisono for directing us to Lemma 16 in \cite{liang_characterizing_2025}.  The authors gratefully acknowledge the mathematical research institute MATRIX in Australia where part of this research was performed.
F.R.C. gratefully acknowledges the ``de Castro" Statistics Initiative at the \textit{Collegio Carlo Alberto} and the \textit{Fondazione Franca e Diego de Castro}. F.R.C. is supported by the Gruppo
Nazionale per l'Analisi Matematica, la Probabilità e le loro Applicazioni (GNAMPA-INdAM).
S.P. gratefully acknowledges funding from UNSW Faculty of Science Research Grant and the Eva Mayr Stihl Foundation.



\bibliographystyle{alpha}
\bibliography{mybibfile.bib}

\appendix

\section{Proofs of variational formulae}

\subsection{Proof of Proposition~\ref{lem:seqsplitpdeWthenFR}}
\label{app:proofseqWFR}
    Here we use the notation $\rho_1$ to refer to the unnormalised form of $\nu_1$.  Denote by $f_\G$ the unnormalised FR operator, 
    $f_\G(\mu) := \mu \log \frac{\pi}{\mu}$.
The unnormalised sequential split solution (in the order W-FR) at time $\gamma$ is given by 
    $\rho_1 = S_\G(\gamma, S_\W(\gamma, \mu_0)) = S_\G(\gamma, \hat{\nu}_1$,
where $S_\G(\gamma, v)$ denotes the abstract solution operator of $\partial_\gamma \rho = f_\G(\rho)$ initialised at $v$ after time $\gamma$ and $S_\W(\gamma,v)$ denotes the solution operator of $\partial_\gamma \rho = f_\W(\rho)$. 
Using the chain rule, 
the first variation of $\rho_1$ with respect to $\gamma$ is  (see also (3.4) in \cite{descombes_lietrotter_2013})
\begin{align}
    \label{eq:derivrhogam}
    \partial_\gamma \rho_1 = f_\G(\rho_1) + \partial_2 S_\G(\gamma, \hat{\nu}_1)f_\W(\hat{\nu}_1), 
\end{align}
where $\hat{\nu}_1 := S_\W(\gamma, \mu_0)$ and $\partial_i$ denotes the directional derivative w.r.t. the $i$-th argument.  Due to \cite[Appendix B.1]{chen2023sampling}, the solution operator $S_\G$ has an explicit form given by $S_\G(\gamma, v) = \pi^{1 - e^{-\gamma}} v^{e^{-\gamma}}$, so that $\partial_2 S_\G(\gamma, v)h$, i.e. the directional derivative w.r.t. the second argument $v \in \cP_2^{ac}(\real^d)$ in the direction $h \in \mathcal{T}$ 
is easily computed as, 
\begin{align*}
    \partial_2 S_\G(\gamma,v)h =  \left. \frac{d}{d\epsilon}S_\G(\gamma, v + \epsilon h)     \right|_{\epsilon = 0} 
    = e^{-\gamma} \pi^{1 - e^{-\gamma}} v^{e^{-\gamma} - 1} h  
    = e^{-\gamma} S_\G(\gamma,v) h v^{-1}. 
\end{align*}
Note that $\partial_2 S_\G(\gamma,v)h $ is a linear operator and  $x\mapsto \partial_2 S_\G(\gamma,v)(x) = e^{-\gamma} (\pi(x)/v(x))^{1-e^{-\gamma}}$ is continuous if $\pi, v$ are continuous and $v(x)>0$.
Substituting back into \eqref{eq:derivrhogam} yields
\begin{align}
    \label{eq:seqsplitpdeWthenFR2}
    \partial_\gamma \rho_1 = f_\G(\rho_1) + e^{-\gamma} S_\G(\gamma,\hat{\nu}_1) f_\W(\hat{\nu}_1) \hat{\nu}_1^{-1} .
\end{align}
Notice that the r.h.s can be written entirely in terms of $\rho_1$ since $S_\G$ is an invertible operator under our assumptions, with inverse given by 
\begin{align*}
     S_\G^{-1}(\gamma,S_\G(\gamma, v))= \pi^{1-e^{\gamma} } S_\G(\gamma, v)^{e^{\gamma}}, 
\end{align*}
Since $\rho_1 = S_\G(\gamma, \hat{\nu}_1)$, we have using the above that,
    $\hat{\nu}_1 = \pi^{1-e^{\gamma} }  \rho_1^{e^{\gamma}}$.
Then $f_\W(\hat{\nu}_1)$ can be expressed in terms of $\rho_1$ since, using shorthand $w := \nabla \log \frac{\rho_1}{\pi}$, 
\begin{align*}
    f_\W(\hat{\nu}_1) &= \nabla \cdot \left(\hat{\nu}_1 \nabla \log \frac{\hat{\nu}_1}{\pi}  \right) \\
     & = e^{\gamma } (\pi^{-e^{\gamma} + 1} \rho_1^{e^{\gamma} - 1} \nabla \cdot \left(\rho_1 w  \right) + \pi^{-e^{\gamma} + 1}\rho_1 w \cdot \nabla  (\rho_1^{e^\gamma - 1})     + \rho_1^{e^\gamma} w \cdot  \nabla (\pi^{-e^{\gamma} + 1} )  ) \\
     & = e^\gamma \pi^{-e^{\gamma} + 1} \rho_1^{e^{\gamma} - 1} f_\W(\rho_1) + e^\gamma (\pi^{-e^{\gamma} + 1}\rho_1 w \cdot \nabla  (\rho_1^{e^\gamma - 1})     + \rho_1^{e^\gamma} w \cdot \nabla (\pi^{-e^{\gamma} + 1} ) ) \\
      & = e^\gamma \pi^{-e^{\gamma} + 1} \rho_1^{e^{\gamma} - 1}  (f_\W(\rho_1) + \underbrace{ \rho_1^{2 - e^{\gamma}}  w \cdot \nabla (\rho_1^{e^{\gamma} - 1}) + \rho_1 \pi^{e^{\gamma}- 1} w \cdot  \nabla  (\pi^{-e^{\gamma} + 1} )  }_{=:R}), 
\end{align*}
and notice that $R$ can be simplified further as 
\begin{align*}
    R & = \rho_1 \left( \rho_1^{1 - e^{\gamma}}  w \cdot \nabla (\rho_1^{e^{\gamma} - 1}) + \pi^{e^{\gamma}- 1} w \cdot \nabla (\pi^{-e^{\gamma} + 1} )  \right) \\
     & = \rho_1 \left( (e^{\gamma} - 1) \rho_1^{-1}  w \cdot  \nabla \rho_1 -  w \cdot (e^{\gamma} - 1) \pi^{-1}\nabla \pi  \right) \\
     & = (e^{\gamma} - 1) \rho_1 |w|^2,  
\end{align*}
so that 
\begin{align*}
    f_\W(\hat{\nu}_1) = e^\gamma \pi^{-e^{\gamma} + 1} \rho_1^{e^{\gamma} - 1}  (f_\W(\rho_1) +  (e^{\gamma} - 1) \rho_1 |w|^2   ) .
\end{align*}
Substituting all the derived expressions back into \eqref{eq:seqsplitpdeWthenFR2} yields 
\begin{align*}
     \partial_\gamma \rho_1 & = f_\G(\rho_1) + e^{-\gamma} e^\gamma \pi^{-e^{\gamma} + 1} \rho_1^{e^{\gamma} - 1}(f_\W(\rho_1) +  (e^{\gamma} - 1) \rho_1 |w|^2   ) \pi^{e^{\gamma} - 1} \cdot \rho_1^{1-e^{\gamma}} \\
     & =  f_\W(\rho_1) + f_\G(\rho_1) +   (e^{\gamma} - 1) \rho_1 |w|^2.
\end{align*}
Then by an application of \cite[Lemma A.1]{us} to obtain the PDE for the normalised density $\nu_1$, the claim holds.     

\subsection{Proof of Proposition~\ref{lem:seqsplitpdeFRthenW}}
\label{app:proofseqFRW}
    By similar reasoning as in Proposition \ref{lem:seqsplitpdeWthenFR}, the sequential split solution at time $\gamma$ in the order FR-W is given by $\eta_1 = S_\W(\gamma, S_\F(\gamma, \mu_0))$ and its first variation w.r.t. $\gamma$ can be computed as 
\begin{align}
	\label{eq:varFRW}
    \partial_\gamma \eta_1 = f_\W(\eta_1) + \partial_2 S_\W(\gamma, \hat{\eta}_1)f_\F(\hat{\eta}_1),
\end{align}
where $\hat{\eta}_1:= S_\F(\gamma, \mu_0)$. By Lemma 1 in \cite{descombes_lietrotter_2013} and recalling that $S_\W$ is a linear operator for which $\partial_2 S_\W(t,v)h = S_\W(t, h)$, we have
\begin{align}
	\nonumber 
     \partial_2 S_\W(\gamma, \hat{\eta}_1)f_\F(\hat{\eta}_1)  &= f_\F(S_\W(\gamma, \hat{\eta}_1)) +  \int_0^\gamma \partial_2 S_\W(\gamma - \tau, \hat{\eta}_1)[f_\W,f_\F](S_\W(\tau, \hat{\eta}_1)) d\tau \\
     \label{eq:pertabs}
     & = f_\F(\eta_1) +  \int_0^\gamma S_\W\left(\gamma - \tau, [f_\W,f_\F](S_\W(\tau, \hat{\eta}_1))\right) d\tau,
\end{align}
since $S_\W$ is a linear operator and where $[f_\W,f_\F](v)$ refers to the Lie commutator (see (3.2a) in \cite{descombes_lietrotter_2013}), 
\begin{align*}
    [f_\W,f_\F](v) &=  D_{\F}f_\W(v) - D_{\W}f_\F(v),  \\
    & = f_\W'f_\F(v) - f_\F'f_\W(v) = v \left( g_1(v)   - \mathbb{E}_{v} \left[  g_1 \right] \right),
\end{align*}
where $D_{\F}$ is the Lie derivative operator associated with $f_\F$, i.e. $D_{\F}H(v)=  H'(v)f_\F(v)$ for any unbounded nonlinear operator $H: X \rightarrow \mathcal{T}$, $X \subset \cP_2^{ac}(\real^d)$ and $H'(v)$ denotes the directional derivative at $v$. 
With a slight abuse of notation, we use $f_\W'f_\F(v)$ in place of $f_\W'(v)f_\F(v)$ to denote the directional derivative of $f_\W$ at $v$ in the direction $f_\F(v)$. 
Substituting \eqref{eq:pertabs} into \eqref{eq:varFRW} yields that the variation of $\eta_1$ takes the form of the usual WFR PDE plus a perturbation term,
\begin{align}
\label{eq:perturbedpde_FRthenW}
    \partial_\gamma \eta_1 = f_\W(\eta_1) + f_\F(\eta_1) + \int_0^\gamma S_\W\left(\gamma - \tau, [f_\W,f_\F](S_\W(\tau, \hat{\eta}_1))\right) d\tau,
\end{align}
where $[f_\W,f_\F]S_\W(\tau, \hat{\eta}_\gamma)\in \mathcal{T}$. 
Substituting in \eqref{eq:commutator1} developed in Lemma~\ref{lem:commut} for the Lie commutator yields \eqref{eq:PDE_splitting2}.

Finally, the perturbation term in~\eqref{eq:perturbedpde_FRthenW}
is characterised in Lemma~\ref{lem:commutator_integral} using the repeated commutators $\mathcal{G}^k$ defined therein, 
\begin{align*}
    \int_0^\gamma S_\W\left(\gamma - \tau, [f_\W,f_\F](S_\W(\tau, \hat{\eta}_1))\right) d\tau =\sum_{k=1}^\infty \frac{\gamma^k}{k!}\mathcal{G}^k(\eta_1).
    \end{align*}
The result then follows using Lemma~\ref{lem:commut}.

\subsection{Auxiliary Lemmas}

We now derive several results that allows us to obtain the variational formulae in Lemma~\ref{lem:seqsplitpdeFRthenW}. The expression in~\eqref{eq:seqsplitpdeFRthenW} is obtained by writing the perturbation term in~\eqref{eq:PDE_splitting2} using the Lie commutator and higher order commutators derived below. 
We work with this formulation as the inverse of the solution operator $S_\W$ does not lead to a tractable form, when it exists. A similar approach can be used to obtain the perturbation term in Proposition~\ref{lem:seqsplitpdeWthenFR} with $g_k = |\nabla g|^2$ for all $k = 1, 2, 3, \dots$. We do not use this approach in the proof of Proposition~\ref{lem:seqsplitpdeWthenFR} but exploit the invertibility of the unnormalised FR solution operator.
In this appendix we rely on the shorthand notation $\eta_\gamma$ to refer to $\eta_1(x;\gamma)$. We start by deriving the commutators.
    \begin{lemma}
    \label{lem:commut} Let Assumption~\ref{ass:lsi} hold. Recall $ \eta_1(x; \gamma)$ denotes the solution of \eqref{eq:seqsplitpdeFRthenW} over a single step of size $\gamma$.  Then the $k$th commutator $ \mathcal{G}^k \eta_1 := [f_\W, \mathcal{G}^{k-1}\eta_1]$ for $k = 1, 2, 3, \dots$ with $\mathcal{G}^0 := f_\F$  has a Fisher-Rao structure, 
    \begin{align*}
        \mathcal{G}^k \eta_1 =\eta_1 \left(g_k(\eta_1) - \mathbb{E}_{\eta_1}[g_k]  \right) 
    \end{align*}
    where $g_{k-1}'f_\W(\eta)$ denotes the directional 
    derivative of $g_{k-1}$ in the direction $f_\W(\eta)$ and 
    \begin{align*}
        g_k(\eta_1) &= -g_{k-1}'f_\W(\eta_1) + \nabla \log \frac{\eta_1}{\pi} \cdot \nabla g_{k-1}(\eta_1) + \frac{1}{\eta_1} \nabla \cdot (\eta_1 \nabla g_{k-1}(\eta_1)), \quad \text{for} \enskip k = 1, 2, 3, \dots \\
                g_0(\eta_1) & =  -  \log \frac{\eta_1}{\pi}
    \end{align*} 
        \end{lemma}
    \begin{proof}
        Once again, with a slight abuse of notation we use $\eta_\gamma$ to denote $\eta_1(x;\gamma)$ to emphasise the dependence on the step size $\gamma$.  We will proceed inductively, starting with $\mathcal{G}^1$, 
\begin{align}
    \label{eq:commWF1}
    [f_\W, f_\F]\eta_\gamma &=  f_\W'f_\F(\eta_\gamma) -f_\F'f_\W(\eta_\gamma).
\end{align}
Since $f_\W$ is a linear operator its directional derivative  
at $\eta_\gamma$ in the direction
$f_\F(\eta_\gamma)$ is $
f_\W(f_\F(\eta_\gamma))$.  Therefore,
\begin{align}
    \label{eq:wassfrechet}
    f_\W'f_\F(\eta_\gamma) &= \Delta \left( \eta_\gamma(g_0 - \mathbb{E}_{\eta_\gamma}[g_0])   \right) - \nabla \cdot \left(\eta_\gamma(g_0 - \mathbb{E}_{\eta_\gamma}[g_0]) \nabla \log \pi  \right) \\
    & =  2 \nabla g_0 \cdot \nabla \eta_\gamma + \eta_\gamma  \Delta g_0 -\eta_\gamma \nabla \log \pi \cdot \nabla g_0  + (g_0 - \mathbb{E}_{\eta_\gamma}[g_0]) f_\W(\eta_\gamma)\notag \\
    &  =  \nabla \cdot (\eta_\gamma \nabla  g_0)  + \left( \eta_\gamma \nabla \log \frac{\eta_\gamma}{\pi} \right) \cdot \nabla g_0  + (g_0 - \mathbb{E}_{\eta_\gamma}[g_0]) f_\W(\eta_\gamma).\notag
\end{align}
Notice that this expression holds for any suitably differentiable scalar valued function $g$ in place of $g_0$.\\

The directional derivative of $f_\F$ is given by 
\begin{align}
 \label{eq:fisherfrechet}
    \frac{d}{d \epsilon}  \left.  f_\F(\eta_\gamma + \epsilon h) \right|_{\epsilon = 0} &=   \frac{d}{d \epsilon}  \left.  -(\eta_\gamma + \epsilon h) \log \frac{\eta_\gamma + \epsilon h}{\pi}  + (\eta_\gamma + \epsilon h)\int  \log \frac{\eta_\gamma + \epsilon h}{\pi} (\eta_\gamma + \epsilon h) 
  \right|_{\epsilon = 0} \\
 & =  \left( \log \frac{\pi}{\eta_\gamma} - 1\right)h - h \int \log \frac{\pi}{\eta_\gamma} \eta_\gamma \,  + \eta_\gamma \int h \left( 1 -  \log \frac{\pi}{\eta_\gamma} \right)  \notag \\
  & = h \left(g_0 - \mathbb{E}_{\eta_\gamma}[g_0] \right) + \eta_\gamma \left(g_0'h - \mathbb{E}_{\eta_\gamma}[g_0'h]  \right) - \eta_\gamma \mathbb{E}_h[g_0]\notag.
\end{align}
Then since $g_0'h = -\frac{h}{\eta_\gamma}$,
\begin{align*}
    f_\F' f_\W(\eta_\gamma) &=   f_\W(\eta_\gamma) \left(g_0 - \mathbb{E}_{\eta_\gamma}[g_0] \right) - \left(f_\W(\eta_\gamma)  - \mathbb{E}_{f_\W(\eta_\gamma)}[1]  \right) - \eta_\gamma \mathbb{E}_{f_\W(\eta_\gamma)}[g_0] \\
    & =   f_\W(\eta_\gamma) \left(g_0 - \mathbb{E}_{\eta_\gamma}[g_0] \right) - \nabla \cdot \left( \eta_\gamma \nabla \log \frac{\eta_\gamma}{\pi} \right)   - \eta_\gamma \mathbb{E}_{f_\W(\eta_\gamma)}[g_0] 
\end{align*}
since $\mathbb{E}_{f_\W(\eta_\gamma)}[1] = 0$.  
Combining yields
\begin{align}
    [f_\W, f_\F]\eta_\gamma 
    & =   f_\W(\eta_\gamma)  + \eta_\gamma \mathbb{E}_{f_\W(\eta_\gamma)}[ g_0]  + \nabla \cdot (\eta_\gamma \nabla  g_0)  +\left( \eta_\gamma \nabla \log \frac{\eta_\gamma}{\pi} \right) \cdot \nabla g_0 \notag\\
    \label{eq:commutator1}
    & = -\eta_\gamma \left( \left| \nabla \log \frac{\eta_\gamma}{\pi} \right|^2   - \mathbb{E}_{\eta_\gamma} \left[  \left| \nabla \log \frac{\eta_\gamma}{\pi} \right|^2 \right] \right)
\end{align}
using the fact that $\mathbb{E}_{f_\W(\eta_\gamma)}[1] = 0$ in the second line and integration by parts in the last line.  This yields the claim of the proof for $k=1$, since by the provided definition of $g_k$, 
\begin{align*}
    g_1 
    & = -g_0'f_\W(\eta_\gamma) - \left|\nabla \log \frac{\eta_\gamma}{\pi} \right|^2 - \frac{1}{\eta_\gamma} \nabla \cdot \left( \eta_\gamma \nabla \log \frac{\eta_\gamma}{\pi} \right) 
     = -\left|\nabla \log \frac{\eta_\gamma}{\pi} \right|^2 .
\end{align*}

Since $\mathcal{G}^1$ has a FR structure, it is immediately clear that $\mathcal{G}^2$ will also, as 
\begin{align*}
    [f_\W, [f_\W, f_\F]] = f_\W'[f_\W, f_\F](\eta_\gamma)-[f_\W, f_\F]'f_\W(\eta_\gamma) , 
\end{align*}
which has the same structure as \eqref{eq:commWF1} with $g_0$ now replaced by $g_1$.  Nevertheless, we will calculate it to show this is the case.  The  
directional derivative of $[f_\W, f_\F]$ is given by
\begin{align*}
    \left. \frac{d}{d \epsilon} [f_\W, f_\F](\eta_\gamma + \epsilon h) \right|_{\epsilon = 0} &= \left. \frac{d}{d \epsilon} -(\eta_\gamma + \epsilon h) \left(\left| \nabla   \log \frac{\eta_\gamma + \epsilon h}{\pi} \right|^2 - \int \left| \nabla   \log \frac{\eta_\gamma + \epsilon h}{\pi} \right|^2 (\eta_\gamma + \epsilon h) \right) \right|_{\epsilon = 0} \\
    & = h (g_1 - \mathbb{E}_{\eta_\gamma}[g_1]) + \eta_\gamma (g_1'h  - \mathbb{E}_{\eta_\gamma}[g_1'h]) - \eta_\gamma \mathbb{E}_{h}[g_1] \notag
\end{align*}
where $g_{1}'h:= -2 \nabla\left(  \frac{h}{\eta_\gamma} \right) \cdot \nabla \log \frac{\eta_\gamma}{\pi}$ is the directional derivative of $g_1$ in the direction $h$. Notice this is the same structure as \eqref{eq:fisherfrechet} but with $g_0$ replaced by $g_1$.  Similarly, we can directly use \eqref{eq:wassfrechet} with $g_1$ in place of $g_0$ for the term $f_\W'[f_\W, f_\F](\eta_\gamma)$ due to the Fisher-Rao structure of $[f_\W, f_\F]$.  This yields
\begin{align*}
    \mathcal{G}^2\eta_\gamma &= f_\W(\eta_\gamma)(g_1 - \mathbb{E}_{\eta_\gamma}[g_1]) + \nabla \cdot (\eta_\gamma \nabla g_1)  + \left( \eta_\gamma \nabla \log \frac{\eta_\gamma}{\pi} \right) \cdot \nabla g_1  \\
    &-\left(f_\W(\eta_\gamma) (g_1 - \mathbb{E}_{\eta_\gamma}[g_1]) +\eta_\gamma (g_1'f_\W(\eta_\gamma)  - \mathbb{E}_{\eta_\gamma}[g_1'f_\W(\eta_\gamma)]) - \eta_\gamma \mathbb{E}_{f_\W(\eta_\gamma)}[g_1] \right)\\
    & = -\eta_\gamma (g_1'f_\W(\eta_\gamma)  - \mathbb{E}_{\eta_\gamma}[g_1'f_\W(\eta_\gamma)]) + \left( \eta_\gamma \nabla \log \frac{\eta_\gamma}{\pi} \right) \cdot \nabla g_1 + \eta_\gamma \mathbb{E}_{f_W(\eta_\gamma)}[g_1]  + \nabla \cdot (\eta_\gamma \nabla g_1).   
\end{align*}
        
Notice that 
\begin{align*}
    \left( \eta_\gamma \nabla \log \frac{\eta_\gamma}{\pi} \right) \cdot \nabla g_1 + \eta_\gamma \mathbb{E}_{f_\W(\eta_\gamma)}[g_1] = \eta_\gamma \left(\nabla \log \frac{\eta_\gamma}{\pi} \cdot \nabla g_1 +\mathbb{E}_{\eta_\gamma} \left[ -\nabla \log \frac{\eta_\gamma}{\pi} \cdot \nabla g_1  \right] \right) 
\end{align*}
using integration by parts for the expectation term. 
For the last term, we can re-write as 
\begin{align*}
    \nabla \cdot (\eta_\gamma \nabla g_1) = \eta_\gamma \left( \frac{1}{\eta_\gamma} \nabla \cdot (\eta_\gamma \nabla g_1) - \mathbb{E}_{\eta_\gamma} \left[\frac{1}{\eta_\gamma} \nabla \cdot (\eta_\gamma \nabla g_1)  \right]    \right) 
\end{align*}
since, by the divergence theorem and the tail decay of $\eta_\gamma, g_1$ guaranteed by Assumption~\ref{ass:lsi}, 
    $\int \nabla \cdot (\eta_\gamma \nabla g_1) = 0$.
Combining the above results yields
\begin{align*}
    \mathcal{G}^2\eta_\gamma &= \eta_\gamma (g_2 - \mathbb{E}_{\eta_\gamma}[g_2]) \\
    g_2 &= -g_1' f_\W(\eta_\gamma) +\nabla \log \frac{\eta_\gamma}{\pi} \cdot \nabla g_1 + \frac{1}{\eta_\gamma} \nabla \cdot (\eta_\gamma \nabla g_1)
\end{align*}
which confirms the claim of the lemma for $k=2$.  By induction, the claim holds for all positive integer $k$.

\end{proof}  

\begin{lemma}
\label{lem:commutator_integral}
    Let Assumption~\ref{ass:lsi} hold.  Recall $ \eta_1(x; \gamma)$ denotes the solution of \eqref{eq:seqsplitpdeFRthenW} over a single step of size $\gamma$.
    The following identity holds formally,
\begin{align*}
    \int_0^\gamma e^{(\gamma - \tau)f_\W} \mathcal{G}^1\eta_1(x;\tau) d\tau= \sum_{k=1}^\infty \frac{\gamma^k}{k!}\mathcal{G}^k(\eta_1(x;\gamma)).
\end{align*}
    \end{lemma}
    \begin{proof}
    Define $f(\tau) := e^{(\gamma - \tau)f_W}\mathcal{G}^1\eta_1(x; \tau)$ and consider a Taylor expansion around point $\gamma$ of $f$,
    $f(\tau) = \sum_{k=0}^\infty \frac{f^{(k)}(\gamma)}{k!}(\tau-\gamma)^k$.
Let us compute the derivatives of $f$.  With a slight abuse of notation, in the remainder of the proof we use $\eta_s$ to denote $\eta_1(x;s)$ to emphasise dependence on $s$. Define also  $\widehat{\eta}_1(x;\gamma):=S_\F(\gamma,\mu_0)$, $\eta_s:=S_\W(s,\widehat{\eta}_1(x;\gamma))$. 
Recall that 
\begin{align*}
    \mathcal{G}^1v =[f_\W,f_\F](v)  = f_\W(f_\F(v)) - f_\F'(v)[f_\W(v)].
\end{align*}
Applying the chain rule we have
    \begin{align*}
        \partial_s e^{(\gamma - s)f_\W}\mathcal{G}^1\eta_s &= -f_\W e^{(\gamma - s)f_\W}\mathcal{G}^1\eta_s + e^{(\gamma - s)f_\W}\partial_s \mathcal{G}^1\eta_s 
    \end{align*}
and by chain rule and linearity of $f_\W$,
\begin{align}
\label{eq:firstintegr}
    \partial_s f_\W(f_\F(\eta_s)) = f_\W'(f_\F(\eta_s))\left[ \partial_s f_\F(\eta_s) \right] = f_\W( f'_\F(\eta_s) [\partial_s \eta_s]) = f_\W( f'_\F(\eta_s) [f_\W(\eta_s)])
\end{align}
since $\partial_s \eta_s = \partial_s S_\W(s, \widehat{\eta}_1(x;\gamma)) = f_\W(\eta_s)$. 
To compute $\partial_s f_\F'(\eta_s)[f_\W(\eta_s)]$ we recall that
\begin{align}
\label{eq:secondfrech}
   \partial_s f_\F'(\eta_s)[f_\W(\eta_s)] &= f_\F''(\eta_s)(\partial_s \eta_s, f_\W(\eta_s)) + f_\F'(\eta_s)[\partial_s f_\W(\eta_s)] \\
    & = f_\F''(\eta_s)(f_\W(\eta_s), f_\W(\eta_s)) + f_\F'(\eta_s)[\partial_s f_\W(\eta_s)] \notag \\
    & = f_\F''(\eta_s)(f_\W(\eta_s), f_\W(\eta_s)) + f_\F'(\eta_s)[f_\W^2(\eta_s)]\notag
\end{align}
where in the last line we have used $\partial_s \eta_s = f_\W(\eta_s)$ and $
    \partial_s^2 \eta_s = \partial_s f_\W(\eta_s) = f_\W'(\eta_s)[\partial_s \eta_s] = f_\W(f_\W(\eta_s)) = f_\W^2(\eta_s)$
by chain rule, and where $ f_\F''(\eta_s)[\partial_s \eta_s, f_\W(\eta_s)]$ is a bilinear map describing how the directional derivative $f_\F'(\eta_s)[f_\W(\eta_s)]$ changes in the direction $\partial_s \eta_s$.

Combining~\eqref{eq:firstintegr}--\eqref{eq:secondfrech} it the immediately follows that
\begin{align*}
\partial_s \mathcal{G}^1\eta_s &= f_\W( f'_\F(\eta_s) [f_\W(\eta_s)])-\left(f_\F''(\eta_s)(f_\W(\eta_s), f_\W(\eta_s)) + f_\F'(\eta_s)[f_\W^2(\eta_s)]\right)\\
&=  f_\W(f_\F'(\eta_s)[f_\W(\eta_s)]) - (f_\F'(\eta_s)[f_\W(\eta_s)])'[f_\W(\eta_s)]\\
&=(\mathcal{G}^1)'(\eta_s) [f_\W(\eta_s)]
\end{align*}
using the chain rule for operators
\begin{align*}
    (g'(v) [f(v)])'[h] = g''(v)[h, f(v)] + g'(v)[f'(v)[h]]
\end{align*}
where $'$ indicates derivative w.r.t. $v$. It immediately follows that
\begin{align*}
    \partial_s \mathcal{G}^1\eta_s  -f_\W(\mathcal{G}^1)(\eta_s)
    &=-[f_\W, [f_\W, f_\F]](S_\W(s, \hat{\eta}_1(x;\gamma)))=-\mathcal{G}^2\eta_s,
\end{align*}
and thus 
\begin{align*}
    f^{(1)}(s) &= -e^{(\gamma-s)f_\W}\mathcal{G}^2\eta_s
\end{align*}
using the fact that $f_\W$ and $e^{t f_\W}$ commute for any $t > 0$. 

For $f^{(2)}$ we have
\begin{align*}
    f^{(2)}(s) &=  f_\W e^{(\gamma - s)f_W}\mathcal{G}^2\eta_s   - e^{(\gamma - s)f_W}\partial_s \mathcal{G}^2\eta_s.
\end{align*}
Recalling that
\begin{align*}
    \mathcal{G}^2v=[f_\W, [f_\W,f_\F]](v)  = f_\W(\mathcal{G}^1v) - (\mathcal{G}^1)'(v)[f_\W(v)],
\end{align*}
we find
\begin{align*}
\partial_{s}f_\W(\mathcal{G}^1\eta_{s})&=f_\W(\mathcal{G}^1\eta_{s})[\partial_{s} \mathcal{G}^1\eta_{s}] =f_\W(\mathcal{G}^1\eta_{s})\left([f_\W, f_\F]'(\eta_{s}) [f_\W(\eta_{s})]\right)\\
\partial_{s}(\mathcal{G}^1)'(\eta_{s})[f_\W(\eta_{s})]&=(\mathcal{G}^1)''(\eta_{s})(f_\W(\eta_{s}), f_\W(\eta_{s}))+(\mathcal{G}^1)'(\eta_{s})[f_\W^2(\eta_{s})].
\end{align*}
Therefore
\begin{align*}
\partial_{s} \mathcal{G}^2\eta_{s} &=f_\W(\mathcal{G}^1\eta_{s})\left([f_\W, f_\F]'(\eta_{s}) [f_\W(\eta_{s})]\right)- (\mathcal{G}^1)''(\eta_{s})(f_\W(\eta_{s}), f_\W(\eta_{s}))+(\mathcal{G}^1)'(\eta_{s})[f_\W^2(\eta_{s})]\\
&=f_\W(\mathcal{G}^1\eta_{s})\left([f_\W, f_\F]'(\eta_{s}) [f_\W(\eta_{s})]\right)-\left((\mathcal{G}^1)'(\eta_{s})f_\W(\eta_{s})\right)'[f_\W(\eta_{s})]\\
&=(\mathcal{G}^2)'(\eta_{s})[f_\W(\eta_{s})]
\end{align*}
using the chain rule for operators.  Following the same steps as before we have
$\partial_{s} \mathcal{G}^2\eta_{s} - f_\W(\mathcal{G}^2\eta_{s}) = -\mathcal{G}^3\eta_{s}$
and thus
    $f^{(2)}(s) = e^{(\gamma - s)f_W}\mathcal{G}^3\eta_{s}$.
A recursive argument then gives 
\begin{align*}
    f^{(k)}(s) &=  (-1)^{k}e^{(\gamma - s)f_W}\mathcal{G}^{k+1}\eta_{s},
\end{align*}
from which follows
\begin{align*}
    e^{(\gamma - \tau) f_\W}\mathcal{G}^1\eta_\tau = \mathcal{G}^1\eta_\gamma +\sum_{k=1}^\infty \frac{(-1)^{k}}{k!}(\tau-\gamma)^k\mathcal{G}^{k+1}\eta_{\gamma}.
\end{align*}
Integrating both sides w.r.t. $\tau$ gives the result.
    \end{proof}

\section{Splitting for multivariate Gaussian distributions}

In this section, $\mu_0(x) = \mathcal{N}(x; m_0, C_0)$ and $\pi(x) = \mathcal{N}(x; m_\pi, C_\pi )$. Also, $b_1, Q_1$ denote mean and covariance of $\nu_1$, the solution of the W-FR scheme after one step and $a_1, P_1$ denote mean and covariance of $\eta_1$, the solution of the FR-W scheme after one step.

\label{sec:AppGauss}

\subsection{Moment ODEs and analytic solutions}

We first refine the result of \cite[Remark 2.3]{liero2025evolution} and obtain an explicit expression for the covariance matrix of the WFR flow rather than its inverse.  The specific form derived below will be particularly useful to obtain the convergence rate of $C_t$ to $C_\pi $ in Lemma~\ref{lem:KLexactWFR}.  Additionally, we obtain explicit solutions for the mean, which are not detailed in \cite{liero2025evolution} and seem to be missing from the literature.  Throughout this section, we focus on PDEs of the form 
\begin{align}
    \label{eq:genPDE}
    \partial_t \mu_t &= f_\W(\mu_t) + f_\F(\mu_t) + \mu_t \left(g_M (\mu_t)  - \mathbb{E}_{\mu_t}[g_M] \right) \\
    g_M(\mu_t) &= \nabla g(\mu_t)^\top M_t \nabla g(\mu_t)
\end{align}
initialised at $\mu_0$ with $g$ as defined in \eqref{eq:geqn} and $M_t \in \mathbb{R}^{d \times d}$ a potentially time dependent symmetric matrix independent of $x, \mu_t$.
The ODEs for the exact WFR are obtained for $M_t\equiv 0$.
Assume further that $C_0$ and $C_\pi$ are invertible (non-degenerate) covariance matrices.  It holds straightforwardly that $\mu_t = \mathcal{N}(x; m_t, C_t)$ and the following lemmas characterise the evolution of $m_t, C_t$ with time.

\begin{lemma}
\label{lem:genmomode}
    \textbf{Moment ODEs, Multivariate Gaussian.} Consider \eqref{eq:genPDE} with the aforementioned conditions.  Then $m_t, C_t$ satisfy for all $t > 0$, 
    \begin{align}
        \label{eq:gencovode}
        \frac{dC_t}{dt} &= -C_t C_\pi^{-1} (I - 2 M_t C_\pi^{-1})C_t  - \left(I -\tfrac{1}{2}C_\pi +2M_t \right)C_\pi^{-1}C_t - C_t C_\pi^{-1} (I -\tfrac{1}{2}C_\pi +2M_t) + 2(I+ M_t)   \\
        \label{eq:genmeanode}
        \frac{dm_t}{dt} 
        & = -(C_tC_\pi^{-1}(I - 2M_t C_\pi^{-1}) + (I + 2 M_t)C_\pi^{-1}   ) (m_t - m_\pi) 
    \end{align}
    \end{lemma}
    \begin{proof}
    By integration and assuming interchange of derivatives and integrals,
    \begin{align}
    \label{eq:covgenPDE}
    \frac{dC_t}{d t} = \int (x - m_t)(x-m_t)^\top \partial_t \mu_t(x) dx  &= \int (x - m_t)(x-m_t)^\top  (f_\W(\mu_t) + f_\F (\mu_t))dx \\
    & + \int (x - m_t)(x-m_t)^\top(g_M(\mu_t) - \mathbb{E}_{\mu_t}[g_M]) \mu_t(x) dx \notag.
\end{align}
The first term on the rhs is obtained directly from the covariance ODE for the exact WFR, see \cite[Appendix D]{us} for a derivation, 
\begin{align}
    \label{eq:exactwfrcovrhs}
    \int (x - m_t)(x-m_t)^\top  (f_\W(\mu_t) + f_\F (\mu_t))dx =-C_t C_\pi^{-1}C_t + C_t -C_\pi^{-1}C_t - C_tC_\pi^{-1} + 2I.
\end{align}
The remaining term is straightforwardly evaluated using that for $\mu_t(x) = \mathcal{N}(x; m_t, C_t)$,  
\begin{align}
    \label{eq:gmexpr}
    g_M(\mu_t) = (x-m_t + m_t - \tilde{m}_t)^\top \tilde{C}_t^{-1}M_t \tilde{C}_t^{-1} (x-m_t + m_t - \tilde{m}_t)
\end{align}
where $\tilde{C}_t^{-1} = C_\pi^{-1} - C_t^{-1}$ and $\tilde{m}_t = \tilde{C}_t (C_\pi^{-1} m_\pi - C_t^{-1} m_t)$.  Then, 
\begin{align}
\nonumber
\int (x  - m_t)(x -m_t)^\top g_M(\mu_t) \mu_t(x)dx 
 = 2(C_t C_\pi^{-1} M_t C_\pi^{-1} C_t - C_t C_\pi^{-1} M_t - M_t C_\pi^{-1} C_t + M_t) \\
+ C_t Tr[C_t \tilde{C}_t^{-1} M_t \tilde{C}_t^{-1}]  + C_t(m_t- \tilde{m}_t)^\top \tilde{C}_t^{-1} M_t \tilde{C}_t^{-1}  (m_t- \tilde{m}_t)
\label{eq:covexp1}
\end{align}
using that for a random variable $X \in \mathbb{R}^{d \times 1}$, $X \sim N(0, \Sigma)$ and any constant matrix $A$, $\mathbb{E}[X X^\top X^\top A X ] = \mathbb{E}[XX^TAXX^T] = \Sigma (A + A^\top) \Sigma + \Sigma Tr[\Sigma A]$. Also, 
\begin{align}
    \label{eq:covexp2}
    &\mathbb{E}_{\mu_t}[g_M] \int (x  - m_t)(x -m_t)^\top  \mu_t(x)dx \\
    &\qquad= C_t \left(Tr[C_t \tilde{C}_t^{-1} M_t \tilde{C}_t^{-1} ]  +  (m_t- \tilde{m}_t)^\top \tilde{C}_t^{-1} M_t \tilde{C}_t^{-1} (m_t- \tilde{m}_t) \right). \notag
\end{align}
Substituting \eqref{eq:covexp1} and \eqref{eq:covexp2} into \eqref{eq:covgenPDE} and combining with \eqref{eq:exactwfrcovrhs} yields the covariance ODE \eqref{eq:gencovode}. The mean ODE is obtained in a similar way and its proof is therefore omitted.

    \end{proof}

\begin{lemma}
\label{lem:analsolngaussian}
    \textbf{Analytic Solutions, Multivariate Gaussian}.  Given the conditions of Lemma \ref{lem:genmomode} 
    and  $\Gamma:= C_\pi^{-1} + \tfrac{1}{2}I$.  Then the mean and covariance ODEs \eqref{eq:genmeanode} and \eqref{eq:gencovode} with initial conditions $m_0, C_0$  have analytic solution for all $t > 0$ given by
    \begin{align}
        \label{eq:gencovanaly}
        C_t &=  C_\pi + e^{-\Gamma t}E_0(I - (W_t^{-1} + E_0)^{-1}E_0) e^{-\Gamma t}  \\ 
        \label{eq:genmeananaly}
        m_t &= m_\pi + (C_t - C_\pi) e^{tC_\pi^{-1}}(C_0 - C_\pi)^{-1}(m_0 - m_\pi).
    \end{align}
    where $E_0:= C_0 - C_\pi$, $W_t:=  (2I + C_\pi )^{-1} -Z_t^M  -e^{-\Gamma t}(2I + C_\pi )^{-1}e^{-\Gamma t}$, $Z_t^M:= 2\int_0^t e^{-\Gamma s} M_s e^{-\Gamma s}ds$.  Assume that $M_s$ is such that $\int_0^t\|e^{-\Gamma s}M_s e^{-\Gamma s}\|\,ds<\infty$ for every finite $s$.
    Notice that $m_t$ is independent of $M_t$.

    An alternative representation for the covariance for the case $C_0 \neq C_\pi$ is given by 
    \begin{align}
    \label{eq:CtintermsofCo_Cpi}
    C_t = C_\pi + (e^{\Gamma t}[(C_0 - C_\pi )^{-1} + (2I + C_\pi )^{-1} -C_\pi^{-1}Z_t^M C_\pi^{-1}]e^{\Gamma t} -(2I + C_\pi )^{-1} )^{-1}.
\end{align}
\end{lemma}
    \begin{proof}
    Beginning with the mean ODE, consider the transformation 
    \begin{align}
        \label{eq:meantransform}
        z_t = (C_t - C_\pi)^{-1}(m_t - m_\pi)
    \end{align}
    and differentiating with respect to $t$ and using \eqref{eq:genmeanode} and $(C_t - C_\pi )z_t = m_t- m_\pi$ yields
    \begin{align*}
        \dot{z_t} 
    & = -(C_t - C_\pi )^{-1} \dot{C}_t z_t - (C_t - C_\pi )^{-1} (C_tC_\pi^{-1}(I - 2M_t C_\pi^{-1}) + (I + 2 M_t)C_\pi^{-1}   ) (C_t - C_\pi )z_t 
\end{align*}
Rearranging \eqref{eq:gencovode} as 
\begin{align*}
    \dot{C}_t 
    & = (C_tC_\pi^{-1} (2M_t C_\pi^{-1} - I) - (I + 2M_t)C_\pi^{-1}    )(C_t - C_\pi)  - (C_t - C_\pi) C_\pi^{-1}
\end{align*}
and substituting into the ODE for $z_t$ yields $\dot{z}_t = C_\pi^{-1} z_t$, 
which has analytic solution $z_t = e^{tC_\pi^{-1}}z_0$.  Inserting \eqref{eq:meantransform} and rearranging yields \eqref{eq:genmeananaly}. 

For the covariance ODE, consider the transformation used in \cite[Remark 2.3]{liero2025evolution}, 
\begin{align}
\label{eq:nattransform}
    A_t = C_t^{-1} = C_\pi^{-1} + e^{-\Gamma t} B_t e^{-\Gamma t}
\end{align}
Differentiating and rearranging yields 
\begin{align*}
\nonumber 
    \dot{A}_t & = C_\pi^{-1}(I - 2M_tC_\pi^{-1}) + A_t\left(I  +2M_t \right)C_\pi^{-1} + C_\pi^{-1}(I + 2M_t)A_t   -A_t -2A_t(I +M_t)A_t \\
    \dot{B}_t  & = -2B_t e^{-\Gamma t} (I + M_t)e^{-\Gamma t} B_t 
\end{align*}
Again as in \cite{liero2025evolution}, setting $D_t = B_t^{-1}$ and differentiating with respect to $t$ yields an ODE whose exact solution is given by 
\begin{align*}
    D_t = D_0 + 2\int_0^t e^{-\Gamma s} (I + M_s)e^{-\Gamma s}ds  = (C_0^{-1} - C_\pi^{-1})^{-1} + Z_t, 
\end{align*}
with $Z_t:= 2\int_0^t e^{-\Gamma s} (I + M_s)e^{-\Gamma s}ds$ and since $D_0 = (C_0^{-1} - C_\pi^{-1})^{-1}$.  Inserting into \eqref{eq:nattransform} yields an analytic solution for $C_t^{-1}$.  Note that for the special case $M_t = 0$, the above coincides with the expression for $C_t^{-1}$ in \cite[Remark 2.3]{liero2025evolution} with $\alpha = \beta = 1$ there.  Note also that $Z_t$ is well defined for all $t > 0$ for the choices of $M_t$ considered in this manuscript, since it is easily checked that the integrand approaches zero asymptotically in $t$.  This representation however is not ideal as it forces the condition $C_0^{-1} \neq C_\pi^{-1}$ to ensure invertibility of their difference. Note also that $C_0 = C_\pi$ is a valid initial condition, wherein $\frac{dC_t}{dt} = 0$ for all $t > 0$ and \eqref{eq:genmeanode} remains well-posed.  We will consider an alternative representation, by using Woodbury matrix identity to obtain 
\begin{align}
\label{eq:Ctwood}
     C_t &=  C_\pi - C_\pi e^{-\Gamma t}(D_0 + Z_t + e^{-2\Gamma t}C_\pi )^{-1}e^{-\Gamma t} C_\pi    =: C_\pi + (R_t^{-1} X_t R_t^{-1})^{-1}  
\end{align}
where $R_t:= C_\pi e^{-\Gamma t}$ and $X_t := (C_\pi^{-1} - C_0^{-1})^{-1} -Z_t - e^{-\Gamma t}C_\pi e^{-\Gamma t}$.  Then let $Z_t = Z_t^M + \Gamma^{-1}(I - e^{-2\Gamma t})$ with $Z_t^M:= 2\int_0^t e^{-\Gamma s} M_s e^{-\Gamma s}ds$.  A repeated application of Woodbury matrix identity on $R_t^{-1} X_t R_t^{-1}$ and using that $C_\pi, \, \Gamma$ and $e^{-\Gamma t}$ commute  yields 
\begin{align*}
    R_t^{-1} X_t R_t^{-1} =  e^{\Gamma t}[(C_0 - C_\pi )^{-1} + (2I + C_\pi )^{-1} -C_\pi^{-1}Z_t^M C_\pi^{-1}]e^{\Gamma t} -(2I + C_\pi )^{-1}. 
\end{align*}
Substituting back into \eqref{eq:Ctwood} yields the result.

    \end{proof}

\subsection{Proof of Lemma \ref{lem:KLexactWFR}}
\label{sec:proofKL}
Due to Gaussianity, we have by a simple algebraic manipulation that 
\begin{align}
\label{eq:kl_gaussian}
     \KL(\nu_n ||\pi) &= \frac{1}{2} \left[ -\log \frac{\det Q_{n}}{\det C_\pi }  + (b_n - m_\pi)^\top C_\pi^{-1}(b_n - m_\pi) + Tr[C_\pi^{-1} Q_{n}] - d \right] \\
     & = \frac{1}{2} \left[ -\log [\det (I + C_\pi^{-1}E_n^\beta)] + (\varepsilon_n^\beta)^\top C_\pi^{-1}\varepsilon_n^\beta + Tr[C_\pi^{-1} E_n^\beta]   \right],\notag
\end{align}
where $E_n^\beta$ and $\varepsilon_n^\beta$ are as defined in \eqref{eq:defncovdiff} and \eqref{eq:defnmeandiff} respectively, $\det (I + C_\pi^{-1}E_n^\beta) > 0$ since $\det(A + B) = \det(A)  \det(I + A^{-1} B)$ for square matrices $A, B$ with $A$ invertible.  A similar expression holds for both $\KL(\mu_{n\gamma}||\pi)$ and $\KL(\eta_{n}||\pi)$ with $Q_{n}$ replaced by $C_{n\gamma}$ and $P_{n}$ respectively. 

Applying the single step covariance equations derived in \eqref{eq:QgammaWthenFR} recursively, one readily obtains the covariance recursion for the W-FR scheme, 
\begin{align}
    \label{eq:WthenFRcov1step}
    Q_{n+1} &= C_\pi  +  (e^{\Gamma \gamma} (Q_{n} - C_\pi )^{-1} e^{\Gamma \gamma} + (e^\gamma - 1)C_\pi^{-1})^{-1}, 
\end{align}
with $\Gamma = C_\pi^{-1} + \frac{1}{2}I$.  Likewise, one obtains for the FR-W scheme, 
\begin{align}
\label{eq:FRthenWcov1step}
    P_{n+1} = C_\pi  + \left(e^{\Gamma \gamma}(P_{n}-C_\pi )^{-1} e^{\Gamma\gamma} + (e^\gamma-1)C_\pi^{-1} e^{2\gamma C_\pi^{-1}}\right)^{-1}
\end{align}
Compare these with the recursion obtained from the solving the WFR PDE as derived in \eqref{eq:WFRGaussexactCt} over time interval $[n\gamma, (n+1)\gamma]$, 
\begin{align}
        \label{eq:WFRcov1step}
        C_{(n+1)\gamma} = C_\pi  + (e^{\Gamma \gamma}[(C_{n\gamma} - C_\pi )^{-1} + (2I + C_\pi )^{-1}]e^{\Gamma \gamma} -(2I + C_\pi )^{-1})^{-1}. 
\end{align} 
Rearranging \eqref{eq:WthenFRcov1step}, \eqref{eq:FRthenWcov1step} and  \eqref{eq:WFRcov1step} yields
\begin{align*}
    E_{n+1}^\beta &=  (e^{\Gamma \gamma} (E_n^\beta)^{-1} e^{\Gamma \gamma} + (e^\gamma - 1)C_\pi^{-1})^{-1} \\
    E_{n+1}^\alpha &=  (e^{\Gamma \gamma} (E_n^\alpha)^{-1} e^{\Gamma \gamma} + (e^\gamma - 1)C_\pi^{-1}e^{2\gamma C_\pi^{-1}})^{-1} \\
    E_{n+1} &= (e^{\Gamma \gamma}[E_n^{-1} + (2I + C_\pi )^{-1}]e^{\Gamma \gamma} -(2I + C_\pi )^{-1})^{-1} 
\end{align*}
respectively, and by induction,
\begin{align}
    \nonumber 
    E_n^\beta &= \left(e^{n \gamma \Gamma} E_0^{-1} e^{n \gamma \Gamma} + (e^\gamma - 1)C_\pi^{-1} \sum_{j=0}^{n-1} e^{2j \gamma \Gamma}  \right)^{-1}  \\
      \label{eq:Enbeta} 
    & = e^{-n \gamma \Gamma}\left(E_0^{-1}  + (1 - e^\gamma) (I - e^{2\gamma \Gamma})^{-1}C_\pi^{-1}(I - e^{-2\gamma n \Gamma})   \right)^{-1} e^{-n \gamma \Gamma} \\
    \label{eq:Enalpha}  
    E_n^\alpha 
    & = e^{-n \gamma \Gamma}\left(E_0^{-1}  + (1 - e^\gamma) (e^{-2\gamma C_\pi^{-1}} - e^{\gamma })^{-1}C_\pi^{-1}(I - e^{-2\gamma n \Gamma})   \right)^{-1} e^{-n \gamma \Gamma} \\   
        \label{eq:Enexpr}
    E_n & = e^{-n \gamma \Gamma} \left(E_0^{-1}  + \frac{1}{2}\Gamma^{-1} C_\pi^{-1}( I - e^{-2n \gamma \Gamma}) \right)^{-1}e^{-n \gamma \Gamma} 
\end{align}
since $\Gamma, e^{\Gamma}, C_\pi $ all commute. 
In \eqref{eq:Enbeta} and \eqref{eq:Enalpha} we have used that  $\sum_{j=0}^{n-1} T^j = (I - T)^{-1}(I-T^n)$ for $(I-T)$ invertible with $T = e^{2 \gamma \Gamma}$. 

The remaining term involving $\varepsilon_n$ is obtained by rearranging \eqref{eq:QgammaWthenFR},  \eqref{eq:meanmultsplitFRthenWexp} and \eqref{eq:WFRGaussexactCt} to obtain 
\begin{align*}
    \varepsilon_n^\beta &= E_n^\beta e^{\gamma C_\pi^{-1}} (E_{n-1}^\beta)^{-1}\varepsilon_{n-1}^\beta; \qquad
    \varepsilon_n^\alpha =  E_n^\alpha e^{\gamma C_\pi^{-1}} (E_{n-1}^\alpha)^{-1}\varepsilon_{n-1}^\alpha; \qquad
    \varepsilon_n = E_n e^{\gamma C_\pi^{-1}}E_{n-1}^{-1}\varepsilon_{n-1}  
\end{align*}
respectively. By induction, one obtains that, 
\begin{align*}
    (\varepsilon_n^\beta)^\top C_\pi^{-1} \varepsilon_n^\beta = \varepsilon_0^\top E_0^{-1} e^{n \gamma C_\pi^{-1}} E_n^\beta C_\pi^{-1} E_n^\beta e^{n \gamma C_\pi^{-1}} E_0^{-1} \varepsilon_0
\end{align*}
and the same expression holds for $(\varepsilon_n^\alpha)^\top C_\pi^{-1} \varepsilon_n^\alpha$ and $(\varepsilon_n)^\top C_\pi^{-1} \varepsilon_n$ with $E_n^\beta$ replaced by $E_n^\alpha$ and $E_n$ respectively. 
Combining all the derived expressions yields the result.

\subsection{Proof of Proposition \ref{prop:KLcomp}}
\label{sec:proofKLsplitvsexact}

\textbf{(i)}
    Here we assume $E_0$ is strictly positive definite and recall $E_n = e^{-n\gamma \Gamma} (E_0^{-1} + \Omega V_n)^{-1} e^{-n\gamma \Gamma}$ where $V_n := C_\pi^{-1} (I - e^{-2n \gamma \Gamma})$.  It trivially holds that $E_n \succ 0$ for all $n = 1, 2, 3 \dots $, recalling that $X Y X$ is p.d. for any $X,Y$ symmetric p.d. It is clear that $e^{-n\gamma \Gamma}$ and $U_n := E_0^{-1} + \Omega V_n$ are both symmetric p.d since $E_0^{-1}$ is the inverse of a symmetric matrix and $\Omega, C_\pi, \Gamma$ are symmetric p.d and they all commute with each other. The same reasoning holds for the claim for $E_n^\beta$ and $E_n^\alpha$.\\
\\
For the second part of the claim, first note that $\Omega, \Omega^\beta, \Omega^\alpha$ are p.d. and 
\begin{align}
\label{eq:omegaordering}
	\Omega^\alpha \succ \Omega \succ \Omega^\beta, 
\end{align}
since $ \Omega^{-1} \prec (\Omega^\beta)^{-1} \Leftrightarrow 2C_\pi^{-1}  \prec \tfrac{e^\gamma}{e^\gamma - 1} (e^{2\gamma C_\pi^{-1}}  - I), $
which is true for any symmetric p.d. $C_\pi$, using the standard matrix exponential inequality, $e^X \succ I + X$ for any symmetric p.d. $X$,  
and that $\tfrac{1}{\gamma} < \frac{1}{1 -e^{-\gamma} }$ for all $\gamma > 0$.  For the first inequality in \eqref{eq:omegaordering}, we have
\begin{align*}
    \Omega^{-1} \succ (\Omega^\alpha)^{-1} & \Leftrightarrow 2C_\pi^{-1} + I \succ \tfrac{1}{1-e^\gamma} (I - e^{2\gamma \Gamma})e^{-2\gamma C_\pi^{-1}} \\
   & \Leftrightarrow 2(e^\gamma -1) C_\pi^{-1}  \succ   I - e^{-2\gamma C_\pi^{-1}} ,  
\end{align*}
which is true for any symmetric p.d. $C_\pi^{-1}$ and $\gamma > 0$ since 
\begin{align*}
   I - e^{-2\gamma C_\pi^{-1}} = 2\gamma C_\pi^{-1} \int_0^1 e^{-2\gamma t C_\pi^{-1}}dt \prec 2\gamma C_\pi^{-1} \prec 2(e^\gamma - 1) C_\pi^{-1}.
\end{align*}
Then since $\Omega^\alpha, \Omega, V_n$ commute and are all p.d.,   
	$(E_0^{-1} + \Omega^\alpha V_n)^{-1}  \prec (E_0^{-1} +  \Omega V_n)^{-1}$.
Also, since $e^{-n\gamma \Gamma} \succ 0$, it holds that 
\begin{align*}
	e^{-n\gamma \Gamma}(E_0^{-1} + \Omega^\alpha V_n)^{-1}e^{-n\gamma \Gamma}  \prec e^{-n\gamma \Gamma}(E_0^{-1} +  \Omega V_n)^{-1}e^{-n\gamma \Gamma}
\end{align*}
that is, $E_n \succ E_n^\alpha \succ 0$.   In particular, since $E_n$ and $E_n^\alpha$ are p.d., this implies that $\lambda_i(E_n) > \lambda_i(E_n^\alpha)$ for all $i = 1, \dots d$ where $\lambda_i(X)$ denotes the $i$th eigenvalue of a matrix $X$. Therefore the claim $||E_n^\alpha||_F < ||E_n||_F$ holds since $||X||_F^2 = \sum_{i=1}^d \lambda_i^2$ for any square matrix $X$ with eigenvalues $\lambda_i$.  By the same arguments, we have that $||E_n^\beta||_F > ||E_n||_F$. 

\textbf{(ii)
}
For the second case, first notice that $E_0 \prec 0$ implies $E_0^{-1} + \Omega C_\pi^{-1} \prec 0$ since 
\begin{align*}
    (C_0 - C_\pi)^{-1} + \Omega C_\pi^{-1} \prec 0  \Leftrightarrow  C_\pi - C_0  \prec C_\pi \Omega^{-1} \Leftrightarrow  C_0 \succ -2I 
\end{align*}
which holds for any p.d $C_0$. Therefore, $
U_n \prec 0$ where $U_n$ is as defined in (i).   Also, for the W-FR split case, $ E_0 \prec 0$ implies $U_n^\beta :=E_0^{-1} + \Omega^\beta V_n \prec 0$ since 
\begin{align*}
    (C_0 - C_\pi)^{-1} + \Omega^\beta C_\pi^{-1}(I - e^{-2n\gamma \Gamma}) \prec 0 & \Leftrightarrow  C_\pi - C_0 \prec (I - e^{-2n\gamma \Gamma})^{-1}C_\pi (\Omega^\beta)^{-1} \\
    & \Leftrightarrow  C_0 \succ C_\pi \left(I - \tfrac{1}{1 - e^\gamma}(I - e^{-2n\gamma \Gamma})^{-1} (I - e^{2\gamma \Gamma}) \right)
\end{align*}
which is true for any p.d $C_0$ since  $\Gamma \succ \tfrac{1}{2}I$ implies $\tfrac{1}{e^\gamma - 1} (e^{2\gamma \Gamma} - I) \succ I - e^{-2n\gamma \Gamma}$
from which it follows that $I - \tfrac{1}{1 - e^\gamma}(I - e^{-2n\gamma \Gamma})^{-1} (I - e^{2\gamma \Gamma})  \prec 0$.  Similarly, for the FR-W Split scheme, $E_0 \prec 0$ implies $U_n^\alpha := E_0^{-1} + \Omega^\alpha V_n \prec 0$ by the same reasoning as for W-FR.  
Then by the same reasoning as in (i), $E_n \prec 0$ and the same holds for  $E_n^\beta$ and $E_n^\alpha$.\\ 
\\
The second part of the claim follows by combining the above and \eqref{eq:omegaordering} to obtain $ 0 \succ U_n^\alpha \succ U_n \succ U_n^\beta,$ from which it follows that $-E_n^\alpha \succ -E_n \succ -E_n^\beta \succ 0,$
and then using the same arguments for the proof of the second claim of (i).

\textbf{(iii)}
We will demonstrate the reasoning for $\KL(\mu_{n \gamma}||\pi)$, noting that the same calculations hold for $\KL(\nu_{n }||\pi)$ and $\KL(\eta_{n }||\pi)$ due to Lemma \ref{lem:KLexactWFR} with $E_n$ replaced by $E_n^\beta$ and $E_n^\alpha$ respectively. 
Using the results of Lemma~\ref{lem:KLexactWFR},
\begin{align}
\label{eq:KLexp1}
            \KL(\mu_{n \gamma}||\pi) &= \frac{1}{2} \left[ -\log \det(I + C_\pi^{-1} E_n) +  Tr[C_\pi^{-1}E_n] + \varepsilon_0^\top E_0^{-1} e^{n \gamma C_\pi^{-1}} E_n C_\pi^{-1} E_n e^{n \gamma C_\pi^{-1}} E_0^{-1} \varepsilon_0  \right].
\end{align}

As $C_\pi^{-1}$ is symmetric p.d, it is orthogonally diagonalisable, i.e. $C_\pi^{-1} = P^\top \Lambda P$ where $P^\top P= I$ and $\lambda_{i} = \Lambda(i,i)$ denoting the $i$th eigenvalue of $C_\pi^{-1}$, $0 < \lambda_1 < \lambda_2 < \lambda_3 \dots < \lambda_d$ and $p_i$, the $i$th column of $P^\top$ is the eigenvector corresponding to $\lambda_i$.  Now, consider the third term in the above, using the expression for $E_n$,
\begin{align*}
    \varepsilon_0^\top E_0^{-1} e^{n \gamma C_\pi^{-1}} E_n C_\pi^{-1} E_n e^{n \gamma C_\pi^{-1}} E_0^{-1} \varepsilon_0 
     &=  e^{-2n\gamma} Tr[P U_n^{-1} E_0^{-1}\varepsilon_0 \varepsilon_0^\top E_0^{-1} U_n^{-1} P^\top e^{-2n\gamma \Lambda}  \Lambda] \\
    & =  \sum_{i=1}^d \lambda_i e^{-2n\gamma (1 + \lambda_i)} N_n(i,i),
\end{align*}
where $U_n$ is as defined in (i) and $N_n:= P U_n^{-1} E_0^{-1}\varepsilon_0 \varepsilon_0^\top E_0^{-1} U_n^{-1} P^\top$. Furthermore, since $C_\pi^{-1}E_n = C_\pi^{-1}C_{n\gamma}-I$ has eigenvalues larger than $-1$, we have $\det(I + C_\pi^{-1}E_n ) = e^{Tr[\log(I + C_\pi^{-1}E_n )]}$, and 
\begin{align*}
    -\log \det (I+ C_\pi^{-1}E_n ) = -Tr[\log (I+C_\pi^{-1}E_n )] = \sum_{k=1}^\infty \frac{(-1)^{k}}{k} Tr[(C_\pi^{-1}E_n)^k] 
\end{align*} 
whenever $||C_\pi^{-1}E_n || < 1$ (so that the series converges), which holds for sufficiently large $n$.  Then the first and second terms in \eqref{eq:KLexp1} are given by  
\begin{align*}
    -\log \det (I + C_\pi^{-1} E_n) + Tr[C_\pi^{-1} E_n] &= \sum_{k=2}^\infty \frac{(-1)^{k}}{k} Tr[(C_\pi^{-1}E_n)^k]  \\
    & =  \sum_{k=2}^\infty \frac{(-1)^{k}}{k} e^{-n\gamma k} \, Tr[( D_n M_n  )^k], 
\end{align*}
where $M_n:= P U_n^{-1} P^\top$ and $D_n: = e^{-n\gamma \Lambda}\Lambda  e^{-n\gamma \Lambda}$ is a diagonal matrix. Starting with $k=2$, 
\begin{align*}
Tr[( D_n M_n)^2] &= \sum_{i=1}^d \sum_{j=1}^d D_n(i,i) D_n(j,j) M_n(i,j)M_n(j,i)  = \sum_{i=1}^d \sum_{j=1}^d \lambda_i \lambda_j e^{-2n\gamma (\lambda_i + \lambda_j)}  M_n(i,j)M_n(j,i), 
\end{align*}
where the terms in the summand decay exponentially with $n$ at rate $2n \gamma (\lambda_i + \lambda_j)$.  A simple induction argument shows that as $k$ increases, the summands decay at increasing rates, since $\lambda_i > 0$.  Combining all and using the same reasoning for $\KL(\nu_n||\pi)$ yields that $\frac{\KL(\nu_n|| \pi)}{\KL (\mu_{n \gamma}|| \pi)}$ is given by 
\begin{align*}
      &\frac{ \sum_{i=1}^d \sum_{j=1}^d \lambda_i \lambda_j e^{-2n \gamma (\lambda_i + \lambda_j + 1)} M_n^\beta(i,j) M_n^\beta(j,i) +  \sum_{i=1}^d \lambda_i e^{-2n\gamma (\lambda_i + 1)} N_n^\beta(i,i) + T_\beta }{\sum_{i=1}^d \sum_{j=1}^d \lambda_i \lambda_j e^{-2n \gamma (\lambda_i + \lambda_j + 1)} M_n(i,j) M_n(j,i) +  \sum_{i=1}^d \lambda_i e^{-2n\gamma (\lambda_i + 1)} N_n(i,i) + T} \\
      &= \frac{\sum_{i=1}^d e^{-2n\gamma (\lambda_i + 1)} \lambda_i N_n^\beta(i,i)  + \mathcal{O}(e^{-2n\gamma (2\lambda_1 + 1)})}{\sum_{i=1}^d e^{-2n\gamma (\lambda_i + 1)} \lambda_i N_n(i,i)  + \mathcal{O}(e^{-2n\gamma (2\lambda_1 + 1)})} \\
    & = \frac{\lambda_1 N_n^\beta(1,1) + \mathcal{O}(e^{-2n\gamma \min(\lambda_1, \lambda_2 - \lambda_1)})}{\lambda_1 N_n(1,1) + \mathcal{O}(e^{-2n\gamma \min(\lambda_1, \lambda_2 - \lambda_1)})} 
\end{align*}
where $T_\beta := \sum_{k=3}^\infty \frac{(-1)^{k}}{k} e^{-n\gamma k}Tr[(D_n M_n^\beta )^k]$ and likewise for $T$ with $M_n^\beta$ replaced by $M_n$ and $M_n^\beta$ and $N_n^\beta$ take the same form as $M_n$ and $N_n$ respectively but with $U_n$ replaced by $U_n^\beta$.  Note that in the third line, we have divided top and bottom by $e^{-2n\gamma (\lambda_1 + 1)}$ and also noting that $U_n \rightarrow E_0^{-1} + \Omega C_\pi^{-1}$ and $U_n^\beta \rightarrow E_0^{-1} + \Omega^\beta C_\pi^{-1}$ as $n \rightarrow \infty$. Taking the limit of the above expression as $n \rightarrow \infty$ yields \eqref{eq:KLsplitratio}.


\section{Negativity of covariance term}
\label{app:covariance}
\begin{lemma}
    Suppose $\nu(x) = \mathcal{N}(x; b, Q)$ and $\pi(x) = \mathcal{N}(x; m_\pi, C_\pi)$ with $C_\pi, Q \succ 0$ satisfying 
    \begin{align}
    \label{eq:covcond}
         \textrm{sym}(C_\pi Q^{-1}) = \frac{1}{2} (C_\pi Q^{-1} + Q^{-1} C_\pi) \succ I. 
    \end{align}
    Then~\eqref{eq:covariance_assumption} holds.

\end{lemma}

\begin{proof}
       Recall that
\begin{align*}
    \Cov_{\tilde{\nu}_\tau} \left( g(\nu), \left|\nabla g(\nu)\right|^2 \right) 
    = \mathbb{E}_{\tilde{\nu}_\tau}  \left[\log \frac{\nu}{\pi} \left|\nabla\log \frac{\nu}{\pi}\right|^2   \right] - \mathbb{E}_{\tilde{\nu}_\tau}  \left[ \log \frac{\nu}{\pi} \right] \mathbb{E}_{\tilde{\nu}_\tau}  \left[  \left|\nabla\log \frac{\nu}{\pi}\right|^2  \right]
\end{align*}
Furthermore, $\tilde{\nu}_\tau = \mathcal{N}(x; b_\tau, Q_\tau)$ since $\tilde{\nu}_\tau$ corresponds to a unit time interpolation between two Gaussians.  Starting with the second term in the above, and using the shorthand notation  $B:= C_\pi^{-1} - Q^{-1}$,  $a:= m_\pi - b_\tau $ and $l := b - b_\tau $, 
\begin{align*}
  \mathbb{E}_{\tilde{\nu}_\tau} \left[ \log \frac{\nu}{\pi} \right] 
  & = \tfrac{1}{2} \left( Tr[BQ_\tau] + c_1 +\log \frac{\det C_\pi}{\det Q}\right) \\
  \mathbb{E}_{\tilde{\nu}_\tau} \left[ \left| \nabla  \log \frac{\nu}{\pi} \right|^2 \right] 
  & = Tr[B^{2}Q_\tau] + c_2 
\end{align*}
where $c_1:= a^\top C_\pi^{-1} a - l^\top Q^{-1}l, c_2:= a^\top C_\pi^{-2} a + l^\top Q^{-2}l -2a^\top C_\pi^{-1} Q^{-1} l$ and  using the identity that if $X \sim \mathcal{N}(m, C)$ then $\mathbb{E}[h(X)] = Tr[PC] + (m-b)^\top P(m-b)$ where $h(X) := (X-b)^\top P(X-b)$.  Also, suppose $X \sim \tilde{\nu}_\tau$ and $Y = X - b_\tau \sim \mathcal{N}(0, Q_\tau)$. Then,
\begin{align*}
      &\mathbb{E}_{\tilde{\nu}_\tau} \left[\log \frac{\nu}{\pi} \left|\nabla\log \frac{\nu}{\pi}\right|^2   \right] 
       = \tfrac{1}{2} \left(\mathbb{E}[Y^\top BY Y^\top B^2 Y] +  \mathbb{E}[Y^\top B Y c_2] + 4\mathbb{E}[Y^\top v_1 Y^\top v_2  ] + \mathbb{E}[c_1 Y^\top B^2 Y] + c_1c_2 \right)  \\
      & = \tfrac{1}{2} \left(Tr[(2Q_\tau B Q_\tau  + Q_\tau  Tr[Q_\tau  B])B^2]  + c_2 Tr[BQ_\tau ] + 4Tr[v_1 v_2^\top Q_\tau]   + c_1 Tr[B^2 Q_\tau] + c_1 c_2 \right)\\
      &+\tfrac{1}{2} \left(\log\frac{\det C_\pi}{\det Q}\left(Tr(B^2Q_\tau)+c_2\right)\right) 
\end{align*}
where $v_1:= C_\pi^{-1}a - Q^{-1}l, v_2 :=  C_\pi^{-2} a + Q^{-2}l - C_\pi^{-1} Q^{-1} l - Q^{-1} C_\pi^{-1} a$ and using that $Y$ has mean zero under $\tilde{\nu}_\tau$ and has zero third moment and using the identity $ \mathbb{E}[Y^\top BYY^\top B^2 Y] =  Tr[(\Sigma  (B + B^\top) \Sigma + \Sigma Tr[\Sigma B])B^2]$ for any $Y \sim \mathcal{N}(0, \Sigma)$.  Combining the above yields 
\begin{align*}
    \Cov_{\tilde{\nu}_\tau} \left( g(\nu), \left|\nabla g(\nu)\right|^2 \right) 
   & = Tr[Q_\tau BQ_\tau B^2] +  2Tr[v_1 v_2^\top Q_\tau] \\
   & = Tr[Q_\tau BQ_\tau B^2] +  2v_1^\top Q_\tau B v_1. 
\end{align*}
As shown in \cite[Lemma E.3]{chen2023sampling}, the covariance of the unit time interpolation between the two Gaussians $\nu, \pi$ is given by 
\begin{align*}
     Q_\tau &= ((1-\tau)Q^{-1} + \tau C_\pi^{-1})^{-1} 
\end{align*}
Now notice that 
 \begin{align*}
     2v_1^\top Q_\tau B v_1 = v_1^\top \left(Q_\tau B + (Q_\tau B)^\top \right)v_1 =  v_1^\top \left(Q_\tau B + B Q_\tau  \right)v_1
 \end{align*} 
 We will obtain conditions ensuring $2v_1^\top Q_\tau B v_1 < 0$ and then show that this condition also implies that $Tr[Q_\tau BQ_\tau B^2] < 0$.  Let $H_\tau:= Q_\tau B + B Q_\tau$, and notice that 
 \begin{align*}
    Q_\tau B = K_\tau^{-1} M
 \end{align*}
 where $M:= QC_\pi^{-1} - I$ and $K_\tau := I  + \tau M$. This then yields 
 \begin{align*}
     K_\tau H_\tau K_\tau^\top &= M + M^\top + 2\tau M M^\top \\
     & = K_1 H_1 K_1 ^\top + 2(\tau - 1)M M^\top. 
 \end{align*}
Clearly $ 2(\tau - 1)M M^\top \preceq 0$ when $\tau \in [0,1]$ and if $H_1 \prec 0$, then $K_1 H_1 K_1 ^\top \prec 0$ by a congruence transformation since $K_1$ is invertible. This then implies $K_\tau H_\tau K_\tau^\top \prec 0$.  That is, a sufficient condition is  that 
 \begin{align*}
     H_1 = 2I - (C_\pi Q^{-1} + Q^{-1} C_\pi) \prec 0 
 \end{align*} 
 from which we immediately obtain condition \eqref{eq:covcond}.  Rearranging this condition yields 
 \begin{align}
    \label{eq:EQcond}
     EQ^{-1} + Q^{-1} E \succ 0
 \end{align}
 where $E:= C_\pi - Q$.  Since $E$ is symmetric, it has an eigendecomposition with real eigenvalues and let $(v_i, \lambda_i)$ denote an eigenvector-eigenvalue pair.  Then,
 \begin{align*}
     v_i^\top(EQ^{-1} + Q^{-1} E)v_i = 2 \lambda_i v_i^\top Q^{-1} v_i > 0.
 \end{align*}
Since $Q^{-1}$ is positive definite, $v_i^\top Q^{-1} v_i > 0$ for all non-zero $v_i \in \mathbb{R}^d$ which therefore implies that $\lambda_i > 0$ for all $i$ and therefore $E$ is positive definite.  This then implies that $B \prec 0$. Then we can rewrite the trace term as 
 \begin{align*}
     Tr[Q_\tau BQ_\tau B^2] = Tr[AD] 
 \end{align*}
 where $A:= Q_\tau^{1/2} B Q_\tau^{1/2}$ and $D:= Q_\tau^{1/2} B^2 Q_\tau^{1/2}$.   Clearly $Q_\tau \succ 0$ and also $A \prec 0$ by congruence transformation another argument.  By similar reasoning, it holds that $D \succ 0$.   Recall that if $A \prec 0$ and $D \succ 0$ with $A,D$ symmetric and of the same size then $Tr[AD] < 0$. 
\end{proof}


\begin{lemma}
\label{app:cov1D}
    Suppose $\nu, \pi\in\cP(\real)$ with $\nu(x) = \nu(2a-x)$ and $\pi(x) = \pi(2a-x)$.
    Let $g(\nu) = \log \frac{\nu(x)}{\pi(x)}$ be such that $\nabla^2 g \preceq \alpha I \preceq 0$.
    Assume additionally that $\nu\neq\pi$ and that neither $h=-g$ nor $|\nabla h|^2$ is constant $\tilde{\nu}_\tau$-almost surely.
      Then~\eqref{eq:covariance_assumption} holds.
\end{lemma}

    \begin{proof}  
It follows from the symmetry of $\nu, \pi
$ that $\tilde{\nu}_\tau(x) \propto  \nu(x)^{(1-\tau)} \pi(x)^\tau$ is also symmetric around $a$ and such that $\int_{x<a} \tilde{\nu}_\tau(x)dx = \int_{x>a} \tilde{\nu}_\tau(x)dx = 1/2$ for all $\tau\in[0, 1]$.

Let us denote $h=-g$, so that $    \text{Cov}_{\tilde{\nu}_\tau} (g, |\nabla g|^2 ) = - \text{Cov}_{\tilde{\nu}_\tau} (h, |\nabla h|^2 )$. Under our assumptions, $h$ and $|\nabla h|^2$ are both decreasing on $\{x<a\}$ and increasing on $\{x>a\}$. 
Since $h$ and $|\nabla h|^2$ are both decreasing in $\{x<a\}$ and both increasing in $\{x>a\}$ we can apply Chebyshev’s integral inequality (e.g. \cite{egozcue2009some})
\begin{align*}
    \int_{\{x<a\}}  h(x)|\nabla h(x)|^2 \tilde{\nu}_\tau(x) dx \geq \frac{1}{\int_{\{x<a\}} \tilde{\nu}_\tau(x) dx} \left( \int_{\{x<a\}} h(x)\tilde{\nu}_\tau(x)dx \right) \left( \int_{\{x<a\}} |\nabla h(x)|^2 \tilde{\nu}_\tau(x)dx  \right) 
\end{align*}
and likewise over $\{x>a\}$.
Then, 
\begin{align*}
    \text{Cov}_{\tilde{\nu}_\tau} (h, |\nabla h|^2 ) 
    & \geq \left(\frac{1}{\int_{\{x<a\}} \tilde{\nu}_\tau(x) dx} - 2 \right) \left( \int_{\{x<a\}} h(x) \tilde{\nu}_\tau(x)dx \right) \left( \int_{\{x<a\}} |\nabla h(x)|^2 \tilde{\nu}_\tau(x)dx  \right) \\
    & + \left(\frac{1}{\int_{\{x>a\}} \tilde{\nu}_\tau(x) dx} - 2\right) \left( \int_{\{x>a\}} h(x) \tilde{\nu}_\tau(x)dx \right) \left( \int_{\{x>a\}} |\nabla h(x)|^2 \tilde{\nu}_\tau(x)dx  \right).
\end{align*}
where in the last line we used the symmetry assumption to obtain $\int_{\{x>a\}} \tilde{\nu}_\tau(x)h(x)dx =\int_{\{x<a\}} \tilde{\nu}_\tau(x)h(x)dx $ and similarly for $ |\nabla h|^2$.
Recalling that $\int_{x<a} \tilde{\nu}_\tau(x)dx = \int_{x>a} \tilde{\nu}_\tau(x)dx = 1/2$ it follows that 
   $ -\text{Cov}_{\tilde{\nu}_\tau} (g, |\nabla g|^2 )=\text{Cov}_{\tilde{\nu}_\tau} (h, |\nabla h|^2 )\geq 0$.
As equality can only be achieved when $h$ or $|\nabla h|^2$ is constant $\tilde{\nu}_\tau$-a.s. the result follows.
\end{proof}

\end{document}